\documentclass[sn-basic, Numbered]{sn-jnl}

\usepackage{graphicx}%
\usepackage{multirow}%
\usepackage{amsmath,amssymb,amsfonts}%
\usepackage{amsthm}%
\usepackage{mathrsfs}%
\usepackage[title]{appendix}%
\usepackage{xcolor}%
\usepackage{textcomp}%
\usepackage{manyfoot}%
\usepackage{booktabs}%
\usepackage{algorithm}%
\usepackage{algorithmicx}%
\usepackage{algpseudocode}%
\usepackage{listings}%

\usepackage{inconsolata}
\usepackage{tikz}
\usepackage{forest}
\usetikzlibrary{trees, positioning, shapes, shadows, arrows.meta}
\usepackage{times}
\usepackage{latexsym}
\usepackage{enumitem}
\usepackage{amsmath,amsfonts,bm}

\def\eqref#1{equation~\ref{#1}}

\def\1{\bm{1}}

\DeclareMathAlphabet{\mathsfit}{\encodingdefault}{\sfdefault}{m}{sl}
\SetMathAlphabet{\mathsfit}{bold}{\encodingdefault}{\sfdefault}{bx}{n}

\def\gA{{\mathcal{A}}}

\def\gD{{\mathcal{D}}}
\def\gE{{\mathcal{E}}}
\def\gF{{\mathcal{F}}}

\def\gR{{\mathcal{R}}}
\def\gS{{\mathcal{S}}}
\def\gT{{\mathcal{T}}}
\def\gU{{\mathcal{U}}}

\def\gY{{\mathcal{Y}}}

\def\sT{{\mathbb{T}}}

\theoremstyle{thmstyleone}%
\theoremstyle{thmstyletwo}%

\theoremstyle{thmstylethree}%

\begin{document}

\title[Article Title]{A Survey on Self-Evolution of Large Language Models}


\author[1,2]{\fnm{Zhengwei} \sur{Tao}}

\author[2]{\fnm{Ting-En} \sur{Lin}}

\author[1]{\fnm{Xiancai} \sur{Chen}}

\author[2]{\fnm{Hangyu} \sur{Li}}

\author[2]{\fnm{Yuchuan} \sur{Wu}}

\author*[2]{\fnm{Yongbin} \sur{Li}}\email{shuide.lyb@alibaba-inc.com}

\author*[1]{\fnm{Zhi} \sur{Jin}}\email{zhijin@pku.edu.cn}

\author[2]{\fnm{Fei} \sur{Huang}}

\author[3]{\fnm{Dacheng} \sur{Tao}}

\author[2]{\fnm{Jingren} \sur{Zhou}}

\affil[1]{\orgdiv{Key Lab of HCST (PKU), MOE}, \orgname{SCS, Peking University}}

\affil[2]{\orgdiv{Alibaba Group}}

\affil[3]{\orgname{Nanyang Technological University}}




\abstract{Large language models (LLMs) have significantly advanced in various fields and intelligent agent applications. However, current LLMs that learn from human or external model supervision are costly and may face performance ceilings as task complexity and diversity increase. 
To address this issue, self-evolution approaches that enable LLM to autonomously acquire, refine, and learn from experiences generated by the model itself are rapidly growing. This new training paradigm inspired by the human experiential learning process offers the potential to scale LLMs towards superintelligence.
In this work, we present a comprehensive survey of self-evolution approaches in LLMs. We first propose a conceptual framework for self-evolution and outline the evolving process as iterative cycles composed of four phases: experience acquisition, experience refinement, updating, and evaluation. 
Second, we categorize the evolution objectives of LLMs and LLM-based agents; then, we summarize the literature and provide taxonomy and insights for each module.
Lastly, we pinpoint existing challenges and propose future directions to improve self-evolution frameworks, equipping researchers with critical insights to fast-track the development of self-evolving LLMs. Our corresponding GitHub repository is available at \href{https://github.com/AlibabaResearch/DAMO-ConvAI/tree/main/Awesome-Self-Evolution-of-LLM}{https://github.com/AlibabaResearch/DAMO-ConvAI/tree/main/Awesome-Self-Evolution-of-LLM}. \textsuperscript{\dag} 
\footnotetext{\textsuperscript{\dag} Work done while Zhengwei Tao was interning at Alibaba Group.}
}


\maketitle

\section{Introduction}

\begin{figure}[!t]
    \includegraphics[width=\textwidth]{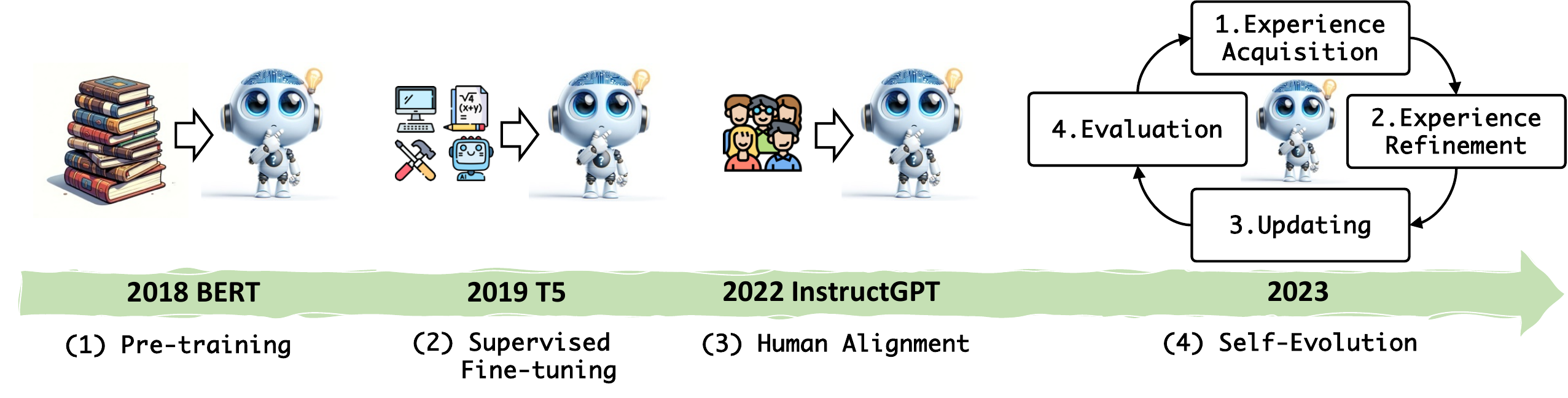}
    \caption{Training paradigms shift of LLMs. }
    \label{fig:intro}
\end{figure}

With the rapid development of artificial intelligence, large language models (LLMs) like GPT-3.5~\cite{ouyang2022training}, GPT-4~\cite{achiam2023gpt}, 
Gemini~\cite{team2023gemini}, LLaMA~\cite{touvron2023llama, touvron2023llama2}, and Qwen~\cite{bai2023qwen} mark a significant shift in language understanding and generation. 
These models undergo three stages of development as shown in Figure~\ref{fig:intro}: pre-training on large and diverse corpora to gain a general understanding of language and world knowledge~\cite{devlin2018bert, brown2020language}, followed by supervised fine-tuning to elicit the abilities of downstream tasks~\cite{raffel2020exploring, chung2022scaling}. Finally, the human preference alignment training enables the LLMs to respond as human behaviors~\cite{ouyang2022training}.
Such successive training paradigms achieve significant breakthroughs, enabling LLMs to perform a wide range of tasks with remarkable zero-shot and in-context capabilities, such as question answering~\cite{tan2023evaluation}, mathematical reasoning~\cite{collins2023evaluating}, code generation~\cite{liu2024your}, and task-solving that require interaction with environments~\cite{liu2023agentbench}.

Despite these advancements, humans anticipate that the emerging generation of LLMs can be tasked with assignments of greater complexity, such as scientific discovery~\cite{miret2024llms} and future events forecasting~\cite{schoenegger2024ai}. However, current LLMs encounter challenges in these sophisticated tasks due to the inherent difficulties in modeling, annotation, and the evaluation associated with existing training paradigms~\cite{burns2023weak}.
Furthermore, the recently developed Llama-3 model has been trained on an extensive corpus comprising 15 trillion tokens\footnote{https://huggingface.co/meta-llama/Meta-Llama-3-70B-Instruct}. It's a monumental volume of data, suggesting that significantly scaling model performance by adding more real-world data could pose a limitation.

This has attracted interest in self-evolving mechanisms for LLMs, akin to the natural evolution of human intelligence and illustrated by AI developments in gaming, such as the transition from AlphaGo~\cite{silver2016mastering} to AlphaZero~\cite{silver2017mastering}. AlphaZero's self-play method, requiring no labeled data, showcases a path forward for LLMs to surpass current limitations and achieve superhuman performance without intensive human supervision. 

Drawing inspiration from the paradigm above, research on the self-evolution of LLMs has rapidly increased at different stages of model development, such as self-instruct \cite{wang2023self}, self-play \cite{tu2024towards}, self-improving \cite{huang2022large}, and self-training \cite{gulcehre2023reinforced}. Notably, DeepMind's AMIE system \cite{tu2024towards} outperforms primary care physicians in diagnostic accuracy, and Microsoft's WizardLM-2 
\footnote{https://wizardlm.github.io/WizardLM2/} exceeds the performance of the initial version of GPT-4. 
Both models are developed using self-evolutionary frameworks with autonomous learning capabilities and represent a potential LLM training paradigm shift.
However, the relationships between these methods remain unclear, lacking systematic organization and analysis.

\begin{figure}[!tb]
    \centering
    \includegraphics[width=1\columnwidth]{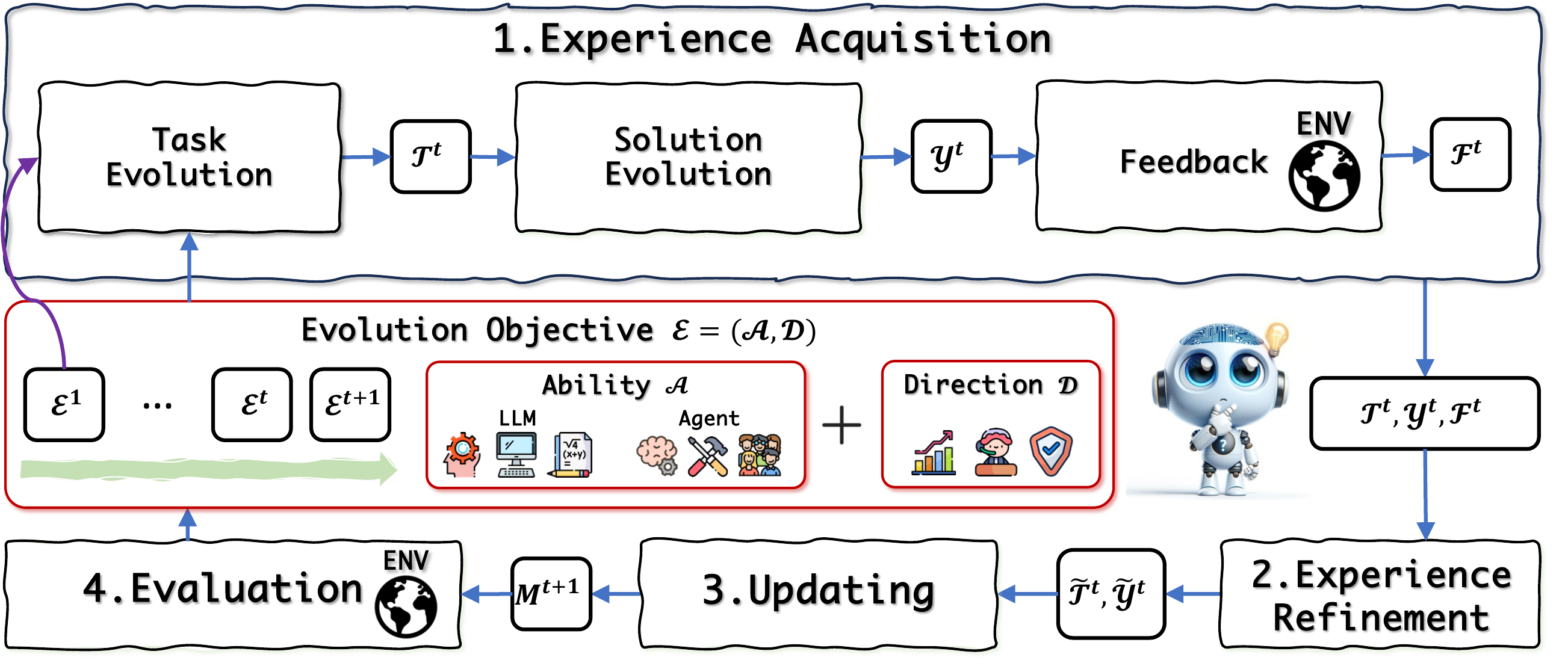}
    \caption{Conceptual framework of self-evolution. For the $t^{th}$ iteration: $\gE^{t}$ is the evolution objective; $\gT^{t}$ and $\gY^{t}$ denote the task and solution; $\gF^{t}$ represents feedback; $M^{t}$ is the current model. Refined experiences are marked as $\Tilde{\gT}^{t}$ and $\Tilde{\gY}^{t}$, leading to the evolved model $\Tilde{M}$. $\mathrm{ENV}$ is the environment. The whole self-evolution starts at $\gE^{1}$.}
    \label{fig:overview}
\end{figure}

Therefore, we first comprehensively investigate the self-evolution processes in LLMs and establish a conceptual framework for their development. This self-evolution is characterized by an iterative cycle involving experience acquisition, experience refinement, updating, and evaluation, as shown in Figure~\ref{fig:overview}. During the cycle, an LLM initially gains experiences through evolving new tasks and generating corresponding solutions, subsequently refining these experiences to obtain better supervision signals. After updating the model in-weight or in-context, the LLM is evaluated to measure progress and set new objectives.


The concept of self-evolution in LLMs has sparked considerable excitement across various research communities, promising a new era of models that can adapt, learn, and improve autonomously, akin to human evolution in response to changing environments and challenges. Self-evolving LLMs are not only able to transcend the limitations of current static, data-bound models but also mark a shift toward more dynamic, robust, and intelligent systems.
This survey deepens understanding of the emerging field of self-evolving LLMs by providing a comprehensive overview through a structured conceptual framework.
We trace the field's evolution from the past to the latest cutting-edge methods and applications while examining existing challenges and outlining future research directions, paving the way for significant advances in developing self-evolution frameworks and next-generation models.


The survey is organized as follows: We first present the overview of self-evolution (\S~\ref{sec:overview}), including background and conceptual framework. 
We summarize existing evolving abilities and domains of current methods(\S~\ref{sec:objective}). 
Then, we provide in-depth analysis and discussion on the latest advancements in different phases of the self-evolution process, including experience acquisition (\S~\ref{sec:gaining experience}), experience refinement (\S~\ref{sec:refining experience}), updating (\S~\ref{sec:updating}), and evaluation (\S~\ref{sec:evaluation}). 
Finally, we outline open problems and prospective future directions (\S~\ref{sec:future}).

\section{Overview}
\label{sec:overview}
In this section, we will first discuss the background of self-evolution and then introduce the proposed conceptual framework.

\subsection{Background}

\noindent\textbf{Self-Evolution in Artificial Intelligence.}
Artificial Intelligence represents an advanced form of intelligent agent, equipped with cognitive faculties and behaviors mirroring those of humans. The aspiration of AI developers lies in enabling AI to harness self-evolutionary capabilities, paralleling the experiential learning processes characteristic of human development.
The concept of self-evolution in AI emerges from the broader fields of machine learning and evolutionary algorithms \cite{back1993overview}. Initially influenced by the principles of natural evolution, such as selection, mutation, and reproduction, researchers have developed algorithms that simulate these processes to optimize solutions to complex problems. The landmark paper by \citet{holland1992adaptation}, which introduced the genetic algorithm, marks a foundational moment in the history of AI's capability for self-evolution. Subsequent developments in neural networks and deep learning have furthered this capability, allowing AI systems to modify their own architectures and improve performance without human intervention \cite{liu2021survey}. 

\noindent\textbf{Can Artificial Entities Evolve Themselves?}
Philosophically, the question of whether artificial entities can self-evolve touches on issues of autonomy, consciousness, and agency. While some philosophers argue that true self-evolution in AI would require some form of consciousness or self-awareness, others maintain that mechanical self-improvement through algorithms does not constitute genuine evolution \cite{chalmers1997conscious}. This debate often references the works of thinkers like  \citet{dennett1993consciousness}, who explore the cognitive processes under human consciousness and contrast them with artificial systems. Ultimately, the philosophical inquiry into AI's capacity for self-evolution remains deeply intertwined with interpretations of what it means to 'evolve' and whether such processes can purely be algorithmic or must involve emergent consciousness \cite{searle1986minds}.


\subsection{Conceptual Framework}
\label{sec:conceptual framework}


In the conceptual framework of self-evolution, we describe a dynamic, iterative process mirroring the human ability to acquire and refine skills and knowledge. This framework is encapsulated within Figure~\ref{fig:overview}, emphasizing the cyclical nature of learning and improvement. Each iteration of the process focuses on a specific evolution goal, allowing the model to engage in relevant tasks, optimize its experiences, update its architecture, and evaluate its progress before moving to the next cycle.

\noindent\paragraph{Experience Acquisition}
At the $t^{th}$ iteration, the model identifies an evolution objective $\gE^{t}$. Guided by this objective, the model embarks on new tasks $\gT^{t}$, generating solutions $\gY^{t}$ and receiving feedback $\gF^{t}$ from the environment, $\mathrm{ENV}$. This stage culminates in the acquisition of new experiences ${(\gT^{t}, \gY^{t}, \gF^{t})}$.

\noindent\paragraph{Experience Refinement}
After experience acquisition, the model examines and refines these experiences. This involves discarding incorrect data and enhancing imperfect ones, resulting in refined outcomes ${(\Tilde{\gT}^{t}, \Tilde{\gY}^{t})}$.

\noindent\paragraph{Updating}
Leveraging the refined experiences, the model undergoes an update process, integrating ${(\Tilde{\gT}^{t}, \Tilde{\gY}^{t})}$ into its framework. This ensures the model remains current and optimized.

\noindent\paragraph{Evaluation}
The cycle concludes with an evaluation phase, where the model's performance is assessed through an evaluation in external environment. The outcomes of this phase inform the objective $\gE^{t+1}$, setting the stage for the subsequent iteration of self-evolution.

\begin{figure}
    \centering
    \tikzset{
        basic/.style  = {draw, text width=3cm, align=center, font=\scriptsize\rmfamily, rectangle, rounded corners=2pt, thin},
        root/.style   = {basic, text width=5cm, fill=green!30, rotate=90},
        node_lv1/.style = {basic, text width=1.7cm, fill=blue!20},
        node_lv1_5/.style = {basic, text width=1.4cm, fill=blue!20},
        node_lv2/.style = {basic, text width=1.8cm, fill=pink!60},
        node_lv3/.style = {basic, text width=11.6cm, fill=yellow!30, align=left},
        node_lv3_5/.style = {basic, text width=9.8cm, fill=yellow!30, align=left},
        edge from parent/.style={draw=black, edge from parent fork right}
    }
    \tiny
    \begin{forest}
        for tree={
            grow=east,
            growth parent anchor=west,
            parent anchor=east,
            child anchor=west,
        }
        [Self-Evolution of LLMs, root, parent anchor=south
            [Evaluation (\S\ref{sec:evaluation}), node_lv1
                [Qualitative, node_lv2,
                    [\citet{yang2023failures} {,} 
                    LLM Explanation \cite{zheng2024judging} {,}
                    ChatEval \cite{chan2023chateval}
                    , node_lv3],
                ]
                [Quantitative, node_lv2, 
                    [
                    LLM-as-a-Judge \cite{zheng2024judging, dubois2024alpacafarm} {,} 
                    Reward Score \cite{ouyang2022training}
                    , node_lv3],
                ]
            ]
            [Updating (\S\ref{sec:updating}), node_lv1,
                [In-Context, node_lv2
                    [\textit{Working Memory}: 
                    Reflexion~\cite{shinn2023reflexion}{,} 
                    IML~\cite{wang2024memory}{,}
                    EvolutionaryAgent \cite{li2024agent} {,}
                    Agent-Pro~\cite{zhang2024agent}{,} 
                    ProAgent~\cite{zhang2024proagent}
                    , node_lv3],
                    [\textit{External Memory}: 
                    MoT~\cite{li2023mot}{,}
                    MemoryBank~\cite{zhong2024memorybank}{,}
                    TiM~\cite{liu2023think}{,}
                    IML~\cite{wang2024memory}{,}
                    TRAN~\cite{yang2023failures}{,}
                    MemGPT~\cite{packer2023memgpt}
                    UA$^2$ \cite{yang2024towards} {,}
                    ICE~\cite{qian2024investigate}{,}
                    AesopAgent~\cite{wang2024aesopagent}
                    , node_lv3],
                ]
                [In-Weight, node_lv2
                    [\textit{Architecture}: LoRA \cite{hu2021lora}{,}
                     ConPET \cite{song2023conpet}{,}
                     Model Soups \cite{wortsman2022model}{,}
                     DAIR \cite{yu2023language}{,}
                     UltraFuser \cite{ding2024mastering}{,}
                     EvoLLM \cite{akiba2024evolutionary}
                     , node_lv3],
                    [\textit{Regularization}: InstuctGPT \cite{ouyang2022training}{,}
                      FuseLLM \cite{wan2024knowledge} {,}
                     Elastic Reset \cite{noukhovitch2024language}{,}
                     WARM \cite{rame2024warm}{,}
                     AMA \cite{lin2024mitigating}
                     , node_lv3],
                    [\textit{Replay}: 
                      ReST \cite{gulcehre2023reinforced, aksitov2023rest} {,}
                      AMIE \cite{tu2024towards}{,}
                     SOTOPIA-$\pi$ \cite{wang2024sotopia}{,}
                     LLM2LLM~\cite{lee2024llm2llm}{,}
                     LTC~\cite{wang2023adapting}{,}
                     A$^3$T \cite{yang2024react}{,}
                     SSR \cite{huang2024mitigating}{,}
                     SDFT \cite{yang2024self}
                     , node_lv3],
                ]
            ]
            [Experience Refinement (\S\ref{sec:refining experience}), node_lv1,
                [Correcting, node_lv2,
                    [\textit{Critique-Free}: 
                     STaR~\cite{zelikman2022star}{,}
                     Self-Debugging~\cite{chen2023teaching}{,}
                     IterRefinement~\cite{chen2023iter}{,}
                     Clinical SV~\cite{gero2023self}
                     , node_lv3],
                    [\textit{Critique-Based}: 
                     Self-Refine~\cite{mad2023refine}{,}
                     CAI~\cite{bai2022constitutional}{,}
                     RCI~\cite{kim2023rci} {,}
                     SELF~\cite{lu2023self}{,}
                     CRITIC~\cite{gou2023critic}{,}
                     SelfEvolve~\cite{jiang2023selfevolve}{,}
                     ISR-LLM~\cite{zhou2023isr}{,}
                     Reflexion \cite{shinn2023reflexion}
                     , node_lv3],
                ]
                [Filtering, node_lv2,
                    [\textit{Metric-Free}: 
                    Self-Consistency~\cite{wang2023SC} {,}
                    LMSI~\cite{huang2022large}{,} 
                    Self-Verification~\cite{weng2023self} {,}
                    CodeT~\cite{chen2022codet}
                    , node_lv3]
                    [\textit{Metric-Based}: 
                    ReST$^{EM}$ \cite{singh2023beyond}{,}
                    AutoAct~\cite{qiao2024autoact}{,} 
                    Self-Talk~\cite{ulmer2024bootstrapping}{,}
                    Self-Instruct~\cite{wang2023self}
                    , node_lv3],
                ]
            ]
            [Experience Acquisition (\S\ref{sec:gaining experience}), node_lv1 ,
                [Feedback (\S\ref{sec:feedback}), node_lv1_5
                    [Environment, node_lv2,
                        [
                        SelfEvolve~\cite{jiang2023selfevolve}{,}
                        Self-Debugging~\cite{chen2023teaching}{,}
                        Reflexion~\cite{shinn2023reflexion}{,}
                        CRITIC~\cite{gou2023critic}{,}
                        RoboCat~\cite{bousmalis2023robocat}{,}
                        SinViG~\cite{xu2024sinvig}{,}
                        SOTOPIA-$\pi$ \cite{wang2024sotopia}
                        , node_lv3_5],
                    ]
                    [Model, node_lv2,
                        [
                        Self-Reward~\cite{yuan2024self}{,}
                        LSX~\cite{stammer2023learning}{,}
                        DLMA~\cite{liu2024direct}{,}
                        SIRLC~\cite{pang2023language}{,}
                        Self-Alignment~\cite{zhang2024self}{,}
                        CAI~\cite{bai2022constitutional}{,}
                        Self-Refine~\cite{mad2023refine}
                        , node_lv3_5],
                    ]
                ]
                [Solution (\S\ref{sec:solution evol}), node_lv1_5
                    [Negative, node_lv2,
                        [\textit{Perturbative}:
                        RLCD~\cite{yang2023rlcd}{,}
                        DLMA~\cite{liu2024direct}{,}
                        Ditto~\cite{lu2024large}
                        , node_lv3_5],
                        [\textit{Contrastive}:
                        Self-Reward~\cite{yuan2024self}{,}
                        SPIN~\cite{chen2024self}{,}
                        GRATH~\cite{chen2024grath}{,}
                        Self-Contrast~\cite{zhang2024selfcontrast}{,} 
                        ETO~\cite{song2024trial}{,}
                        A\textsuperscript{3}T~\cite{yang2024react}{,}
                        STE~\cite{wang2024llms}{,}
                        COTERRORSET~\cite{tong2024can}
                        , node_lv3_5],
                    ]
                    [Positive, node_lv2,
                        [\textit{Grounded}:
                        Self-Align~\cite{sun2024principle}{,}
                        SALMON~\cite{sun2023salmon}{,}
                        MemoryBank~\cite{zhong2024memorybank}{,}
                        TiM~\cite{liu2023think}{,}
                        MoT~\cite{li2023mot}{,}
                        IML~\cite{wang2024memory}{,}
                        TRAN~\cite{yang2023failures}{,}
                        MemGPT~\cite{packer2023memgpt}
                        , node_lv3_5
                        ],
                        [\textit{Self-Play}: 
                        Debates~\cite{taubenfeld2024systematic}{,}
                        Self-Talk~\cite{ulmer2024bootstrapping}{,}
                        Ditto~\cite{lu2024large}{,}
                        SOLID~\cite{askari2024self}{,}
                        SOTOPIA-$\pi$ \cite{wang2024sotopia}
                        , node_lv3_5],
                        [\textit{Interactive}: 
                        SelfEvolve~\cite{jiang2023selfevolve}{,} 
                        LDB~\cite{zhong2024ldb}{,} 
                        ETO~\cite{song2024trial}{,}
                        A$^3$T \cite{yang2024react}{,}
                        AutoAct~\cite{qiao2024autoact}{,}
                        KnowAgent~\cite{zhu2024knowagent}
                        , node_lv3_5],
                        [\textit{Rationale-Based}: 
                        LMSI~\cite{huang2022large}{,}
                        STaR~\cite{zelikman2022star}{,} 
                        A$^3$T \cite{yang2024react}
                        , node_lv3_5],
                    ]
                ]
                [Task (\S\ref{sec:task evol}), node_lv1_5
                    [Selective, node_lv2,
                        [
                        DIVERSE-EVOL~\cite{wu2023self}{,}
                        SOFT~\cite{wang2024step}{,}
                        Selective Reflection-Tuning~\cite{li2024selective}{,}
                        V-STaR~\cite{hosseini2024v}
                        , node_lv3_5],
                    ]
                    [Knowledge-Free, node_lv2,
                        [
                        Self-Instruct~\cite{wang2023self, honovich2022unnatural, roziere2023code}{,}
                        Ada-Instruct~\cite{cui2023ada}{,}
                        Evol-Instruct~\cite{xu2023wizardlm}{,}
                        MetaMath~\cite{yu2023metamath}{,}
                        PromptBreeder~\cite{fernando2023promptbreeder}{,} 
                        Backtranslation~\cite{li2023self}{,}
                        Kun~\cite{zheng2024kun}
                        , node_lv3_5],
                    ]
                    [Knowledge-Based, node_lv2,
                        [\textit{Unstructured}: 
                        UltraChat~\cite{ding2023enhancing}{,}
                        SciGLM~\cite{zhang2024sciglm}{,}
                        EvIT~\cite{tao2024evit}{,}
                        MEEL~\cite{tao2024meel}
                        , node_lv3_5],
                        [\textit{Structured}: 
                        Self-Align~\cite{sun2024principle}{,} 
                        Ditto~\cite{lu2024large}{,} 
                        SOLID~\cite{askari2024self}
                        , node_lv3_5],
                    ]
                ]
            ]
        ]
    \end{forest}
    \caption{Taxonomy of self-evolving large language models.}
    \label{fig:lit_surv}
\end{figure}
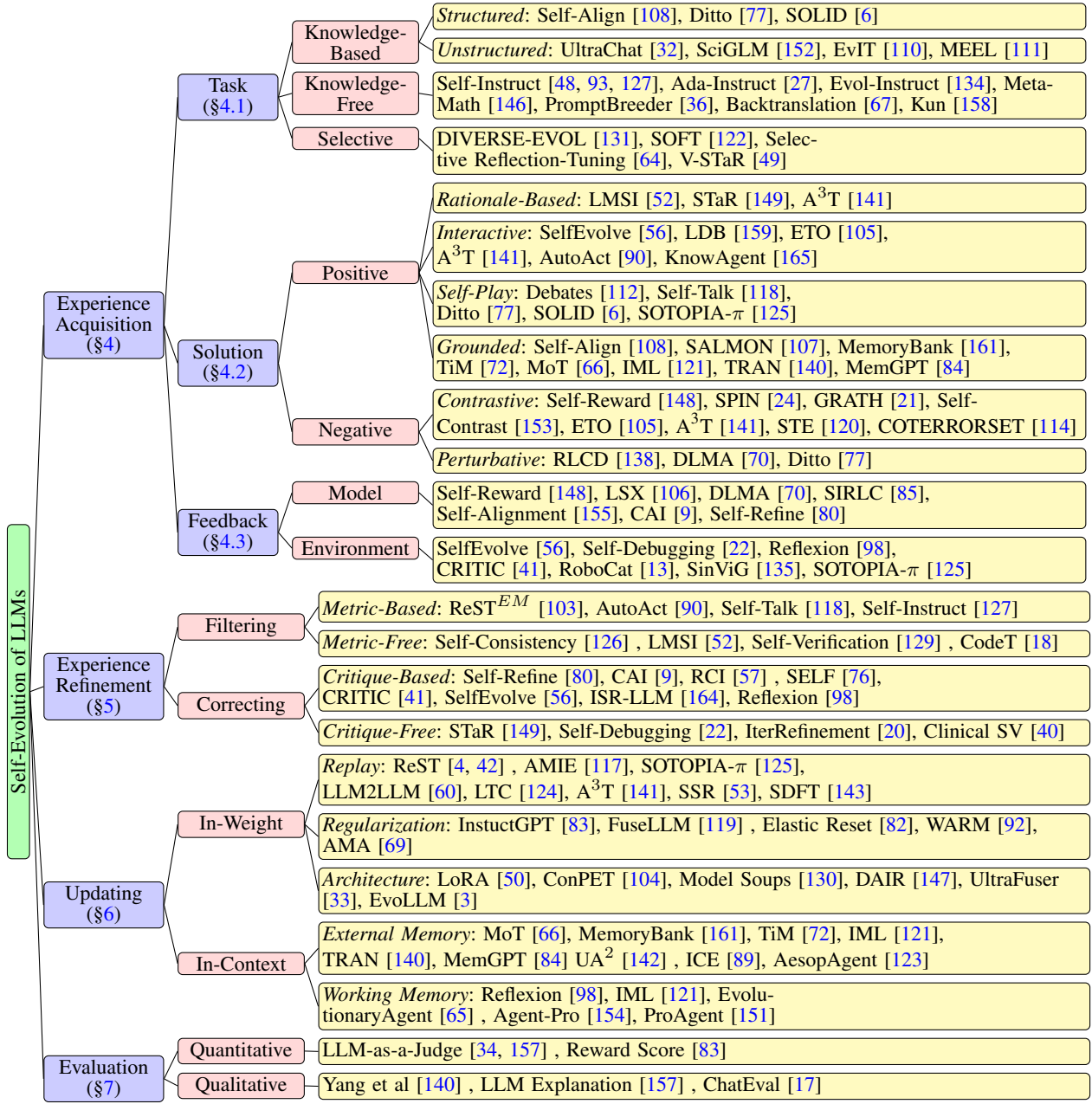

The conceptual framework outlines the self-evolution of LLMs, akin to human-like acquisition, refinement, and autonomous learning processes. We illustrate our taxonomy in Figure~\ref{fig:lit_surv}.

\section{Evolution Objectives}
\label{sec:objective}
Evolution objectives in self-evolving LLMs serve as predefined goals that autonomously guide their development and refinement. Much like humans set personal objectives based on needs and desires, these objectives are crucial as they determine how the model iteratively self-updates. They enable the LLM to autonomously learn from new data, optimize algorithms, and adapt to changing environments, effectively "feeling" its needs from feedback or self-assessment and setting its own goals to enhance functionality without human intervention. 

We define an evolution objective as combining an evolving ability and an evolution direction. An evolving ability stands for an innate and detailed skill. The evolution direction is the aspect of the evolution objective aiming to improve. We formulate the evolution objective as follows:

\begin{equation}
    \gE^{t} = ( \gA^{t}, \gD^{t} ), 
\end{equation}
where $ \gE^{t}$ is the evolution objective, composed by evolving abilities $\gA^{t}$ and  evolution directions $\gD^{t}$. Take "reasoning accuracy improving" as an example, "reasoning" is the evolving ability and "accuracy improving" is the evolution direction.

\subsection{Evolving Abilities}

\begin{table}
\centering 
\setlength{\tabcolsep}{1mm}
\scriptsize
\begin{tabular}{ccccccc}
\toprule 
    \multirow{2}{*}{\textsc{Method}} & \multicolumn{3}{c}{\textsc{Acquisition}} & \multirow{2}{*}{\textsc{Refinement} $f^{\gR}$} & \multirow{2}{*}{\textsc{Updating} $f^{\gU}$} & \multirow{2}{*}{\textsc{Objective} $\gE$} \\
\cmidrule{2-4}
 & \textsc{Task} $f^{\gT}$ & \textsc{Solution} $f^{\gY}$ & \textsc{Feedback} $f^{\gF}$ & & & \\
\midrule 
\multicolumn{7}{c}{\textsc{Large Language Models}}\\
\midrule
    Self-Align~\cite{sun2024principle}    & Knowledge-Based  & Pos-G    & -          & Filtering   & In-W       & \textcolor{brown}{IF}       \\
SciGLM~\cite{zhang2024sciglm}         & Knowledge-Based  & -        & -          & -           & In-W       & \textcolor{blue}{Other}             \\
EvIT~\cite{tao2024evit}               & Knowledge-Based  & -        & -          & -           & In-W       & \textcolor{blue}{Reasoning}        \\
MEEL~\cite{tao2024meel}               & Knowledge-Based  & -        & -          & -           & In-W       & Reasoning         \\
UltraChat~\cite{ding2023enhancing}& Knowledge-Based & -    & -          & -           & In-W       & \textcolor{blue}{Role-Play}   \\
SOLID~\cite{askari2024self}           & Knowledge-Based  & Pos-S    & -          & Filtering   & In-W       & \textcolor{teal}{Role-Play}         \\
Ditto~\cite{lu2024large}              & Knowledge-Based  & Pos-S, Neg-P & -    & -           & In-W       & \textcolor{teal}{Role-Play}         \\
MetaMath~\cite{yu2023metamath}        & Knowledge-Free   & Pos-R    & -          & -           & In-W       & Math              \\
Self-Rewarding~\cite{yuan2024self}    & Knowledge-Free   & -        & Model      & -           & In-W       & \textcolor{teal}{IF,Reasoning,Role-Play}                \\
Kun~\cite{zheng2024kun}               & Knowledge-Free   & -        & -          & Filtering   & In-W       & \textcolor{teal}{IF,Reasoning}                \\
PromptBreeder~\cite{fernando2023promptbreeder} & Knowledge-Free & -    & -          & -           & In-C       & Math, Reasoning   \\
Ada-Instruct~\cite{cui2023ada}        & Knowledge-Free   & -        & -          & -           & In-W       & Math, Reasoning, Code \\
Backtranslation~\cite{li2023self}     & Knowledge-Free   & -        & -          & -           & In-W       & \textcolor{teal}{IF}                \\
DiverseEvol~\cite{wu2023self}         & Selective      & -    & -          & -           & In-W       & Code              \\
Grath~\cite{chen2024grath}            & Selective      & Neg-C    & Model      & -           & In-W       & Reasoning         \\
REST$^{em}$~\cite{singh2023beyond}    & Selective      & -        & Model      & Filtering   & In-W       & Math, Code        \\
SOFT~\cite{wang2024step}              & Selective      & -        & -          & -           & In-W       & \textcolor{brown}{IF}        \\
LSX~\cite{stammer2023learning}        & -          & Pos-R    & Model      & Correcting  & In-W       & Other             \\
LMSI~\cite{huang2022large}            & -          & Pos-R    & -          & Filtering   & In-W       & Math              \\
TRAN~\cite{yang2023failures}          & -              & Pos-G    & -          & -           & In-C       & Reasoning         \\
MOT~\cite{li2023mot}                  & -          & Pos-R, Pos-G & -    & Filtering   & In-C       & Math, Reasoning   \\
STaR~\cite{zelikman2022star}          & -          & Pos-R, Neg-C & Model & Correct     & In-W       & Reasoning         \\
COTERRORSET~\cite{tong2024can}        & -          & Pos-R, Neg-C & -    & -           & In-W       & Math, Reasoning   \\
Self-Debugging~\cite{chen2023teaching} & -            & Pos-I    & Env        & -           & In-C       & Code              \\
SelfEvolve~\cite{jiang2023selfevolve} & -          & Pos-I    & Env          & -           & In-C       & Code              \\
Reflexion~\cite{shinn2023reflexion}   & -          & Pos-I, Pos-G & -    & -           & In-C       & Code, Reasoning   \\
V-STaR~\cite{hosseini2024v}           & -              & Neg-C    & Model      & Filter      & In-W       & Math, Code        \\
Self-Contrast~\cite{zhang2024self}    & -              & Neg-C    & Model      & -           & In-W       & Reasoning         \\
SALMON~\cite{sun2023salmon}           & -              & Neg-C    & Model      & -           & In-W       & \textcolor{teal}{IF,Reasoning,Role-Play}                \\
SPIN~\cite{chen2024self}              & -              & Neg-C    & -          & -           & In-W       & \textcolor{teal}{IF,Reasoning,Role-Play}                \\
RLCD~\cite{yang2023rlcd}              & -              & Neg-P    & Model      & -           & In-W       & \textcolor{brown}{IF}        \\
DLMA~\cite{liu2024direct}             & -              & Neg-P    & Model      & -           & In-W       & \textcolor{brown}{IF}            \\
SELF~\cite{lu2023self}                & -          & -        & Model      & Correct     & In-W       & \textcolor{teal}{IF, Math}          \\
\midrule
\multicolumn{7}{c}{\textsc{LLM Agents}} \\
\midrule
AutoAct~\cite{qiao2024autoact} & Knowledge-Based & Pos-I & Env & Filtering & In-W & Planning, Tool \\
KnowAgent~\cite{zhu2024knowagent} & Knowledge-Based & Pos-I, Pos-G & Env & Filtering & In-W & Embodied, Planning, Tool \\
RoboCat~\cite{bousmalis2023robocat} & Knowledge-Free & Pos-I & Env & - & In-W & Embodied \\
STE~\cite{wang2024llms} & Knowledge-Free & Pos-I, Neg-C & Env & Correct & In-W & Tool \\
IML~\cite{wang2024memory} & - & Pos-R, Pos-G & - & - & In-C & Reasoning \\
SinViG~\citet{xu2024sinvig} & - & Pos-I & Env & Filtering & In-W & Embodied \\
ETO~\cite{song2024trial} & - & Pos-I, Neg-C & Env & Correct & In-W & Tool \\
A\textsuperscript{3}T~\cite{yang2024react} & - & Pos-I, Neg-C & Env & Correct & In-W & Tool \\
Debates~\cite{taubenfeld2024systematic} & - & Pos-S & - & - & In-W & Communication \\
SOTOPIA-$\pi$ \cite{wang2024sotopia} & - & Pos-S,Pos-G & Env & - & In-W & Communication \\
Self-Talk~\cite{ulmer2024bootstrapping} & - & Pos-S, Pos-G & Model & Filtering & In-W & Communication \\
MemGPT~\cite{packer2023memgpt} & - & Pos-G & Env & Filtering & In-C & Communication \\
MemoryBank~\cite{zhong2024memorybank} & - & Pos-G & Env & Filtering & In-C & Communication \\
ProAgent~\cite{zhang2024proagent} & - & Pos-G & Env & - & In-C & Embodied \\
Agent-Pro~\cite{zhang2024agent} & - & Pos-G & Env & - & In-C & Planning \\
AesopAgent~\cite{wang2024aesopagent} & - & Pos-G & Env & - & In-C & Planning \\
ICE~\cite{qian2024investigate} & - & Pos-G & Env & - & In-C & Planning \\
TiM~\cite{liu2023think} & - & Pos-G & - & - & In-C & Communication \\
Werewolf \cite{xu2023exploring} & - & Pos-G & - & - & In-C & Planning \\
\bottomrule 
\end{tabular}
\caption{Overview of self-evolution methods, detailing approaches across evolutionary stages. Key: Pos (Positive), Neg (Negative), R (Rationale-based), I (Interactive), S (Self-play), G (Grounded), C (Contrastive), P (Perturbative), Env (Environment), In-W (In-Weight), In-C (In-Context), IF (Instruction-Following). We use colors to highlight evolution directions: \textcolor{teal}{Adaptation to Feedback} in green, \textcolor{blue}{Expansion of Knowledge Base} in blue, and \textcolor{brown}{Safety, Ethics, and Bias Reduction} in brown. Performance enhancement remains in black.
}
\label{tab:comparison} 
\end{table}

In Table~\ref{tab:comparison}, we summarize and categorize the targeted evolving abilities in current self-evolution research into two groups: LLMs and LLM Agents.

\subsubsection{LLMs}
These are fundamental abilities underlying a broad spectrum of downstream tasks. 

\noindent\textbf{Instruction Following}: The capability to follow instructions is essential for effectively applying language models. It allows these models to address specific user needs across different tasks and domains, aligning their responses within the given context \cite{xu2023wizardlm}.

\noindent\textbf{Reasoning}: LLMs can self-evolve to recognize statistical patterns, making logical connections and deductions based on the information. They evolve to perform better reasoning involving methodically dissecting problems in a logical sequence~\cite{cui2023ada}.

\noindent\textbf{Math}: LLMs enhance the intricate ability to solve mathematical problems covering arithmetic, math word, geometry, and automated theorem proving~\cite{ahn2024large} towards self-evolution. 

\noindent\textbf{Coding}: Methods improve the LLM coding abilities to generate more precise and robust programs~\cite{singh2023beyond, zelikman2023self}. Furthermore, EvoCodeBench~\cite{li2024evocodebench} provides an evolving benchmark that updates periodically to prevent data leakage.

\noindent\textbf{Role-Play}: 
It involves an agent understanding and acting out a particular role within a given context. This is crucial in scenarios where the model must fit into a social structure or follow a set of behaviors associated with a specific identity or function~\cite{lu2024large}. 

\noindent\textbf{Others}: Apart from the above fundamental evolution objectives, self-evolution can also achieve and a wide range of NLP tasks~\cite{stammer2023learning, koa2024learning, gulcehre2023reinforced, zhang2024sciglm, zhang2024selfcontrast}.

\subsubsection{LLM-based Agents}
The abilities discussed here are characteristic of advanced artificial agents used for task-solving or simulations in digital or physical world. These capabilities mirror human cognitive functions, allowing these agents to perform complex tasks and interact effectively in dynamic environments.

\noindent\textbf{Planning}:
It involves the ability to strategize and prepare for future actions or goals. An agent with this skill can analyze the current state, predict the outcomes of potential actions, and create a sequence of steps to achieve a specific objective~\cite{qiao2024autoact}. 

\noindent\textbf{Tool Use}:
This is the capacity to employ objects or instruments in the environment to perform tasks, manipulate surroundings, or solve problems~\cite{zhu2024knowagent}. 

\noindent\textbf{Embodied Control}:
It refers to an agent's ability to manage and coordinate its physical form within an environment. This encompasses locomotion, dexterity, and the manipulation of objects~\cite{bousmalis2023robocat}.

\noindent\textbf{Communication}:
It is the skill to convey information and understand messages from other agents or humans. Agents with advanced communication abilities can participate in dialogue, collaborate with others, and adjust their behaviour based on the communication received~\cite{ulmer2024bootstrapping}.

\subsection{Evolution Directions}
Examples of evolution directions include but are not limited to:

\noindent\textbf{Improving Performance}: The goal is to continuously enhance the model's understanding and generation across various languages and abilities. 
For instance, a model initially trained for question answering and chitchat can autonomously extend its proficiency and develop abilities like diagnostic dialogue \cite{tu2024towards}, social skills \cite{wang2024sotopia}, and role-playing \cite{lu2024large}.


\noindent\textbf{Adaptation to Feedback}: This involves improving model responses based on feedback to better align with preferences or adapt to environments \cite{yang2023rlcd, sun2024principle}. 

\noindent\textbf{Expansion of Knowledge Base}: The aim is to continuously update the model's knowledge base with the latest information and trends. For example, a model might automatically integrate new scientific research into its responses \cite{wu2024continual}.

\noindent\textbf{Safety, Ethic and Reducing Bias}: The goal is to identify and mitigate biases in the model's responses, ensuring fairness and safety. One effective strategy is to incorporate guidelines, such as constitutions or specific rules, to identify inappropriate or biased responses and correct them through model updates \cite{bai2022constitutional, lu2024sofa}.

\section{Experience Acquisition}
\label{sec:gaining experience}
Exploration and exploitation~\cite{gupta2006interplay} are fundamental strategies for learning in humans and LLMs. 
Among that, exploration involves seeking new experiences to achieve objectives and is analogous to the initial phase of LLM self-evolution, known as experience acquisition. This process is crucial for self-evolution, enabling the model to autonomously tackle core challenges such as adapting to new tasks, overcoming knowledge limitations, and enhancing solution effectiveness.
Furthermore, experience is a holistic construct, encompassing not only the tasks encountered~\cite{dewey1938experience} but also the solutions developed to address these tasks~\cite{schon2017reflective}, and the feedback~\cite{boud2013reflection} received as a result of task performance.

Inspired by that, we divide experience acquisition into three parts: task evolution, solution evolution, and obtaining feedback. In task evolution, LLMs curate and evolve new tasks aligning with evolution objectives. For solution evolution, LLMs develop and implement strategies to complete these tasks. Finally, LLMs may optionally collect feedback from interacting with the environment for further improvements.

\subsection{Task Evolution}
\label{sec:task evol}

\begin{figure}[!tb]
    \centering
    \includegraphics[width=1\columnwidth]{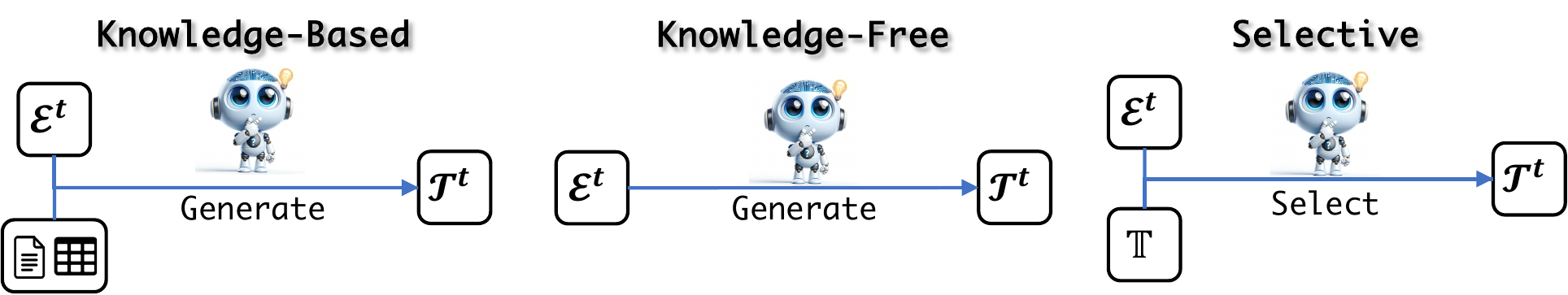}
    \caption{Task evolution. $\gE^{t}$ and $\gT^{t}$ are the evolving objective and task of $t^{th}$ iteration. $\sT$ is the set of all tasks to be selected. The first two are generative methods that differ based on their respective use of knowledge. The third method, in contrast, employs a discriminative approach to select what to learn.}
    \label{fig:task-evolution}
\end{figure}

To gain new experience, the model first evolves new tasks according to the evolution objective $\gE^{t}$ in the current iteration. Task evolution is the crucial step in the engine that starts the entire evolution process. Formally, we denote the task evolution as: 

\begin{equation}
    \gT^{t} = f^{\gT}(\gE^{t}, \mathrm{M}^{t}),
\end{equation}
where $f^{\gT}$ is the task evolution function. $\gE^{t}$, $M^{t}$, and $\gT^{t}$ denote the evolution objective, the model, and the evolved task at iteration $t$, respectively. We summarize and categorize existing studies on the task evolution method $f^{\gT}$ into three groups: \textit{Knowledge-Based}, \textit{Knowledge-Free}, and \textit{Selective}. 
We detail each type in the following parts and show the concepts in Figure~\ref{fig:task-evolution}.

\noindent\paragraph{Knowledge-Based}
The objective $\gE^{t}$ may associate with external knowledge to evolve where the knowledge is not inherently comprised in the current LLMs. 
Explicitly sourcing from knowledge enriches the relevance between tasks and evolution objectives.
It also ensures the validity of relevant facts in the tasks.
We delve into the Knowledge-Based methods seeking to evolve new tasks of the evolving objective assisted by external information.

The first kind of knowledge is structured. Structured knowledge is dense in information and well-organized. 
Self-Align~\cite{sun2024principle} guides task generation by covering 20 topics, such as scientific and legal expertise, to ensure diversity. 
Apart from topic knowledge, DITTO~\cite{lu2024large} includes character knowledge from Wikidata and Wikipedia. The knowledge comprises attributes, profiles, and concise character details for role-play conversations.
SOLID~\cite{askari2024self} generates structured entity knowledge as conversation starters.

The second group consists of tasks evolving from an unstructured context. Unstructured context is easy to obtain but is sparse in knowledge.
UltraChat~\cite{ding2023enhancing} gathers unstructured knowledge of 20 types of text materials to construct questions or instructions. 
SciGLM~\cite{zhang2024sciglm} derives questions from the text of diversified science subjects, which covers rich scientific knowledge.
EvIT~\cite{tao2024evit} derives event reasoning tasks based on large-scale unstructured events mined from the unsupervised corpus.
Similarly, MEEL~\cite{tao2024meel} evolves multi-modal events in both image and text to construct the tasks for MM event reasoning.



\noindent\paragraph{Knowledge-Free}
Unlike previous methods that require extensive human effort to gather external knowledge, Knowledge-Free approaches operate independently using the evolving object $\gE^{t}$ and the model itself. These efficient methods can generate more diversified tasks without additional knowledge restrictions.


First, the LLMs can prompt themselves to generate new tasks according to $\gE^{t}$.
Self-Instruct~\cite{wang2023self, honovich2022unnatural, roziere2023code} is a typical methodology of Knowledge-Free task evolution. These methods self-generate a variety of new task instructions based on evolution objectives.
Ada-Instruct~\cite{cui2023ada} further proposes an adaptive task instruction generation strategy that fine-tunes open-source LLMs to generate lengthy and complex task instructions for code completion and mathematical reasoning. 

Second, extending and boosting original tasks increases the quality of instructions.
WizardLM~\cite{xu2023wizardlm} proposes Evol-Instruct that evolves instruction following tasks with in-depth and in-breadth evolving and further expands it in code generation~\cite{luo2024wizardcoder}.
MetaMath~\cite{yu2023metamath} rewrites the question in multiple ways, including rephrasing, self-verification, and FOBAR. It evolves a new MetaMathQA dataset for fine-tuning LLMs to improve mathematical task-solving.
Promptbreeder~\cite{fernando2023promptbreeder} evolves seed tasks via mutation prompts. It further evolves mutation prompts via the hyper mutation prompts to increase the task diversity.

Third, deriving tasks from plain text is another way. 
Backtranslation~\cite{li2023self} extracts self-contained segments in unlabelled data and regards it as the answers to tasks. 
Similarly, Kun~\cite{zheng2024kun} presents a task self-evolving algorithm utilizing instruction harnessed from unlabelled data towards back-translation. 


\noindent\paragraph{Selective}
Instead of task generation, we may start with large-scale existing tasks. At each iteration, LLMs can select tasks that exhibit the highest relevance to the current evolving objective $\mathcal{E}^{t}$ without additional generation.
This approach obviates the intricate curation of new tasks, streamlining the evolution process~\cite{zhou2024lima, li2023quantity, chen2023alpagasus}.


A simple task selecting method is to randomly sample tasks from the task pool like REST~\cite{gulcehre2023reinforced}, REST$^{em}$~\cite{singh2023beyond}, and GRATH~\cite{chen2024grath} do.
Rather than random selection, DIVERSE-EVOL~\cite{wu2023self} introduces a data sampling technique where the model selects new data points based on their distinctiveness in the embedding space, ensuring diversity enhancement in the chosen subset. SOFT~\cite{wang2024step} then splits the initial training set. Each iteration selects one chunk of the split set as the evolving task. 

\citet{li2024selective} propose Selective Reflection-Tuning and select a subset of tasks via a novel metric calculating to what extent the answer is related to the question.
V-STaR~\cite{hosseini2024v} selects the correct solutions in the previous iteration and adds their task instructions to the task set of the next iteration.

\subsection{Solution Evolution}
\label{sec:solution evol}

\begin{figure}[!tb]
    \centering
    \includegraphics[width=1\columnwidth]{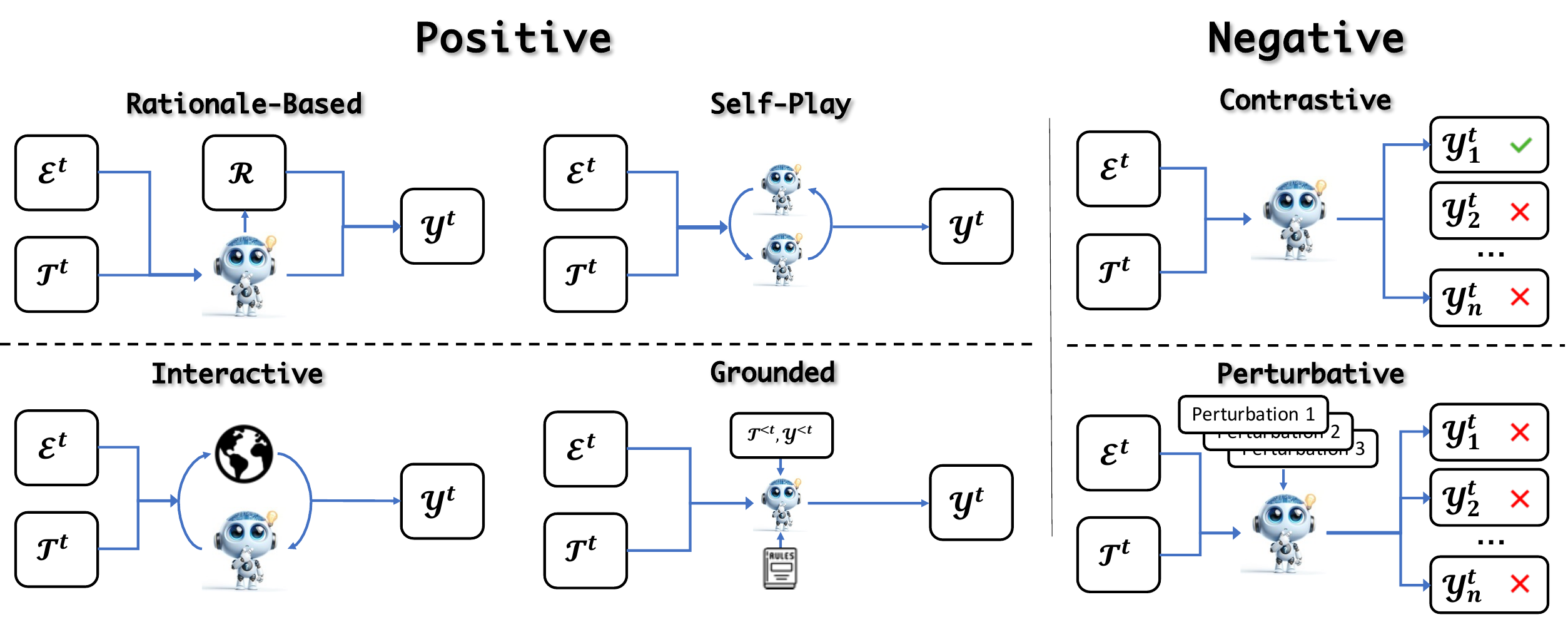}
    \caption{Solution evolution. $\gE^{t}$, $\gT^{t}$, and $\gY^{t}$ are the evolving objective, task, and solution of $t^{th}$ iteration. $\gR$ is the rational thought.}
    \label{fig:solution-evolution}
\end{figure}

After obtaining evolved tasks, LLMs solve the tasks to acquire the corresponding solution. 
The most common strategy is to generate the solution directly according to the task formulation~\cite{zelikman2022star,gulcehre2023reinforced,singh2023beyond,zheng2024kun,yuan2024self}.
However, this straightforward approach might reach solutions irrelevant to the evolution objective, leading to suboptimal evolution~\cite{hare2019dealing}.
Therefore, solution evolution uses different strategies to solve tasks and enhance LLM capabilities by ensuring that solutions are not just generated but are also relevant and informative. 
In this section, we comprehensively survey these strategies and illustrate them in Figure~\ref{fig:solution-evolution}.
We first formulate the solution-evolution as follows:

\begin{equation}
    \gY^{t} = f^{\gY}(\gT^{t}, \gE^{t}, \mathrm{M}^{t}),
\end{equation}
where $f^{\gY}$ is the model's strategy to approach the evolution objective. 

We then categorize these methods into positive and negative according to the correctness of the solutions.
The positive methods introduce various approaches to acquire correct and desirable solutions.
On the contrary, negative methods elicit and collect undesired solutions, including unfaithful or mis-align model behaviors, which are then used for preference alignment.
We elaborate on the details of each type in the following sections.

\subsubsection{Positive}
Current studies explore diverse methods beyond vanilla inference for positive solutions to obtain correct solutions aligned with evolution objectives.
We categorize the task-solving process into four types: Rationale-Based, Interactive, Self-Play, and Grounded.

\noindent\paragraph{Rationale-Based}
The model incorporates rationale explanations towards approaching the evolving objective when solving the tasks and can self-evolve by utilizing such rationales.
These methods enable models to explicitly acknowledge the evolution objective and complete this task in that direction~\cite{wei2022chain, yao2024tree, besta2024graph, yao2022react}. 

\citet{huang2022large} proposes a method where an LLM self-evolves using "high-confidence" rationale-augmented answers generated for unlabeled questions. 
Similarly, STaR~\cite{zelikman2022star} generates rationale when solving the task. If the answer is wrong, it further corrects the rationale and the answer. Then, it uses the answer and rationale as experiences to fine-tune the model.
Similarly, LSX~\cite{stammer2023learning} proposes the novel paradigm to generate an explanation of the answer, incorporating an iterative loop between a learner module performing a base task and a critic module that assesses the quality of explanations given by the learner. 
\citet{song2024trial, yang2024react} obtain rationales in the ReAct~\cite{yao2022react} style when solving the tasks. The rationales are further engaged in training the agents in the following step.

\noindent\paragraph{Interactive}
Models can interact with the environment to enhance the evolution process.
These methods can obtain environmental feedback that is valuable for guiding self-evolution directions. 

SelfEvolve and LDB~\cite{jiang2023selfevolve, zhong2024ldb} improve code generation ability via self-evolution. They allow the model to generate code and acquire feedback via running the code on the interpreter. 
As another environment, \citet{song2024trial, yang2024react} interact in embodied scenarios and acquire feedback. They learn to take proper actions based on their current state.
For agent abilities, AutoAct~\cite{qiao2024autoact} introduces self-planning from scratch, focusing on an intrinsic self-learning process. In this process, agents enhance their abilities through recursive planning iterations with environment feedback.
Following AutoAct, \cite{zhu2024knowagent} further enhances agent training by integrating self-evolution and an external action knowledge base. This approach guides action generation and boosts planning ability through environment-driven corrective feedback loops.

\noindent\paragraph{Self-Play} 
It's the situation where a model learns to evolve by playing against copies of itself.
Self-play is a powerful evolving method because it enables systems to communicate with themselves to get feedback in a closed loop. It's especially effective in environments where the model can simulate various sides of the roles, like multi-player games~\cite{silver2016mastering, silver2017mastering}.
Compared with interactive methods, self-play is an effective strategy to obtain feedback without an environment. 

\citet{taubenfeld2024systematic} investigate the systematic biases in simulations of debates by LLMs. 
On the contrary to debating, \citet{ulmer2024bootstrapping} engage LLMs in conversations following generated principles. 
Another kind of conversation via role-playing. \citet{lu2024large} proposes self-simulated role-play conversation. The process involves instructing the LLM with character profiles and aligning its responses to maintain consistency with the character’s knowledge and style. 
Similarly, \citet{askari2024self} propose SOLID to generate large-scale intent-aware role-play dialogues. This self-playing aspect harnesses the expansive knowledge of LLMs to construct information-rich exchanges that streamline the dialog generation process.
\citet{wang2024sotopia} introduces a novel approach whereby each LLM follows a role and communicates with each other to achieve their goals.

\noindent\paragraph{Grounded}
To reach the evolving objective and reduce exploration space, models can be grounded on existing rules~\cite{sun2024principle} and previous experiences for further explicit guidance when solving the tasks.

LLMs can generate desirable solutions more effectively by being grounded on pre-defined rules and principles.
For instance, Self-Align~\cite{sun2024principle} generated self-evolved questions with principle-driven constraints to guide the task-solving process.
SALMON~\cite{sun2023salmon} design a set of combined principles that requires the model to follow when solving the task.
Self-Talk~\cite{ulmer2024bootstrapping} ensures the LLMs generate a workflow-aligned conversation based on preset agent characters. They generate the workflow in advance based on GPT-4.

Besides pre-defined rules, grounding on previous experiences can improve the solutions.
MemoryBank~\cite{zhong2024memorybank} and TiM~\cite{liu2023think} answer current questions by incorporating previous question-answer records.
Rather than previous solution histories, MoT~\cite{li2023mot}, IML~\cite{wang2024memory}, and TRAN~\cite{yang2023failures} incorporate induced rules from the histories to answer new questions. 
MemGPT~\cite{packer2023memgpt} combines these merits and retrieves previous questions, solutions, induced events, and user portrait knowledge.


\subsubsection{Negative}
In addition to acquiring positive solutions, recent research illustrates that LLMs can benefit from negative ones for self-improvement~\cite{yang2023failures}.
This strategy is analogous to trial and error in human behavior when learning skills. This section summarises typical methods of gaining negative solutions to assist in self-evolution. 

\noindent\paragraph{Contrastive} 
A widely used group of methods is to collect multiple solutions for a task and then contrast the positive and negative ones to get improvements.


For instance, Self-Reward~\cite{yuan2024self} generates responses, predicts corresponding rewards by the model itself, and constructs preference pairs based on rewards. SPIN~\cite{chen2024self} uses the results generated by the current model as negatives and the ground truth of SFT as positives for iterative self-training.
Similarly, GRATH~\cite{chen2024grath} generates both correct and incorrect answers. It then trains the model by comparing these two answers.
Self-Contrast~\cite{zhang2024selfcontrast} contrasts the differences and summarizes these discrepancies into a checklist that could be used to re-examine and eliminate discrepancies.
In ETO~\cite{song2024trial}, the model interacts with the embodied environment to complete tasks and optimizes from the failure solutions.
A\textsuperscript{3}T~\cite{yang2024react} improves ETO by adding rationale after each action for solving tasks. 
STE~\cite{wang2024llms} implements trial and error where the model solves the tasks with unfamiliar tools. It learns by analyzing failed attempts to improve problem-solving strategies in future tasks.
More recently, COTERRORSET~\cite{tong2024can} obtains incorrect solutions generated by PALM-2 and proposes mistake tuning, which requires the model to void making mistakes.
 

\noindent\paragraph{Perturbative}
Compared to \textit{Contrastive}, \textit{Perturbative} methods seek to add perturbations to obtain negative solutions intentionally.
Models can later learn to avoid generating these negative answers.
Adding perturbations to obtain negative solutions is more controllable than contrastive methods. 

Some methods add perturbation to generate harmful solutions~\cite{yang2023rlcd, liu2024direct}.
Given a task, RLCD~\cite{yang2023rlcd} curates both positive and negative instructions and generates positive and negative solutions.
DLMA~\cite{liu2024direct} gathers both positive and negative instructional prompts and subsequently produces corresponding positive and negative solutions.

Rather than harmful perturbation, incorporating negative context is another way.
Ditto~\cite{lu2024large} adds negative persona characters to generate incorrect conversations. The model then learns from the negative conversations to evolve persona dialogue ability.

\subsection{Feedback}
\label{sec:feedback}

\begin{figure}[!tb]
    \centering
    \includegraphics[width=1\columnwidth]{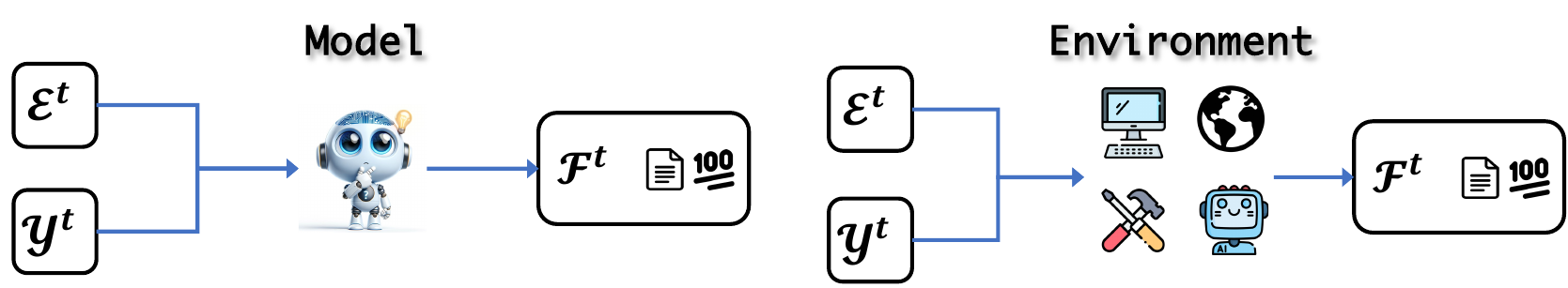}
    \caption{Types of feedback. $\gE^{t}$ and $\gY^{t}$ are the evolving objective and task solutions of $t^{th}$ iteration.}
    \label{fig:feedback}
\end{figure}

As humans learn skills, feedback plays a critical role in demonstrating the correctness of the solutions. This key information enables humans to reflect and then update their skills. 
Akin to this process, LLMs should obtain feedback during or after the task solution in the cycle of self-evolution. We formalize the process as follows:
\begin{equation}
    \gF^{t} = f^{\gF}(\gT^{t}, \gY^{t}, \gE^{t}, \mathrm{M}^{t}; \mathrm{ENV}),
\end{equation}
where $f^{\gF}$ is the method to acquire feedback.

In this part, we summarize two types of feedback. Model feedback refers to gathering the critique or score rated by the LLMs themselves. Besides, Environment denotes the feedback received directly from the external environment. We illustrate these concepts in Figure~\ref{fig:feedback}.

\subsubsection{Model}
Current studies demonstrate that LLMs can play well as a critic~\cite{zheng2024judging}. In the cycle of self-evolution, the model judges itself to acquire the feedback of the solutions.

One type of feedback is a score that indicates correctness.
Self-Reward~\cite{yuan2024self}, LSX~\citet{stammer2023learning}, and DLMA~\cite{liu2024direct} rate their own solutions and output the scores via LLM-as-a-Judge prompting.
Similar to that, SIRLC~\cite{pang2023language} utilizes self-evaluation results of LLM as the reward for further reinforcement learning. 
Self-Alignment~\cite{zhang2024self} leverages the self-evaluation capability of an LLM to generate confidence scores on the factual accuracy of its outputs. 

Another type provides a textual description, offering multi-dimensional information.
To alter the distribution of the responses via supervised learning, CAI~\cite{bai2022constitutional} asks the model to critique its response according to a principle in the constitution. 
In contrast to supervised learning and reinforcement learning approaches, Self-Refine~\cite{mad2023refine} allows the model to generate natural language feedback on its own output in a few-shot manner.

\subsubsection{Environment} 
Another form of feedback comes from the environment, common in tasks where solutions can be directly evaluated. 
This feedback is precise and elaborate and can provide sufficient information for model updating.
They may be derived from code interpreter~\cite{jiang2023selfevolve, chen2023teaching, shinn2023reflexion}, tool execution~\cite{qiao2024autoact, gou2023critic}, the embodied environment~\cite{bousmalis2023robocat, xu2024sinvig, zhou2023isr}, and other LLMs or agents~\cite{wang2024sotopia, taubenfeld2024systematic, ulmer2024bootstrapping}.

For code generation, Self-Debugging~\citet{chen2023teaching} utilizes execution results on test cases as part of feedback while SelfEvolve~\cite{jiang2023selfevolve} receives the error message from the interpreter.
Similarly, Reflexion~\cite{shinn2023reflexion} also obtains the run-time feedback from the code interpreter. It then further reflects to generate thoughts.
This run-time feedback contains the trace-back information that can point out the key information for improved code generation.

Recently, methods endow tool-using ability to LLMs and agents. Executing tools leading to feedback in return~\cite{gou2023critic, qiao2024autoact, song2024trial, yang2024react, wang2024llms}.

RoboCat~\cite{bousmalis2023robocat} and SinViG~\cite{xu2024sinvig} act in the robotic embodied environment. This type of feedback is precise and strong to guide self-evolution.

Communication feedback is common and effective in LLM-based multi-agent systems. Agents can correct and support each other, enabling co-evluation~\cite{wang2024sotopia, taubenfeld2024systematic, ulmer2024bootstrapping}.







\section{Experience Refinement}
\label{sec:refining experience}

After experience acquisition and before updating in self-evolution, LLMs may improve the quality and reliability of their outputs through experience refinement. It helps LLMs adapt to new information and contexts without relying on external resources, leading to more reliable and effective assistance in dynamic environments. This process is formulated as follows:
\begin{equation}
    \Tilde{\gT}^{t}, \Tilde{\gY}^{t} = f^{\gR}(\gT^{t}, \gY^{t}, \gF^{t}, \gE^{t}, \mathrm{M}^{t}),
\end{equation}
where $f^{\gR}$ is the methods of experience refinement, $\Tilde{\gT}^{t}, \Tilde{\gY}^{t}$ are the refined tasks and solutions. We classify the methods into two categories: filtering and correcting.

\begin{figure}
    \centering
    \includegraphics[width=1\columnwidth]{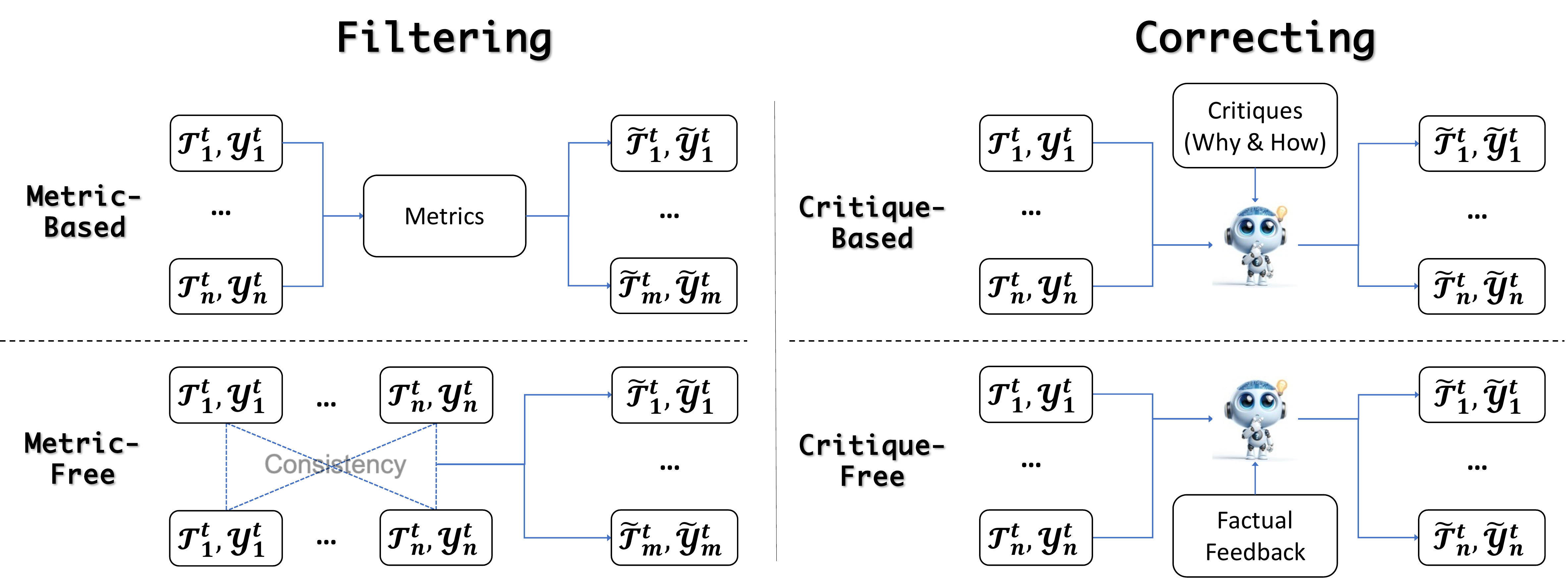}
    \caption{Experience refinement. $\Tilde{\gT}^{t}$ and $\Tilde{\gY}^{t}$ are the task and solution after refinement, respectively. Filtering reduces the number of experiences from $n$ to $m$.}
    \label{fig:exp-refine}
\end{figure}

\subsection{Filtering} 
Refinement in self-evolution involves two primary filtering strategies: \textit{Metric-Based} and \textit{Metric-Free}. The former uses external metrics to assess and filter outputs, while the latter does not rely on these metrics. This ensures that only the most reliable and high-quality data is utilized for further updating.

\subsubsection{Metric-Based}
By relying on feedback and pre-defined criteria, metric-based filtering improves the quality of the outputs~\cite{singh2023beyond, qiao2024autoact, ulmer2024bootstrapping, wang2023self}, ensuring the progressive enhancement of LLM capabilities through each iteration of refinement. 

For example, ReST$^{EM}$~\cite{singh2023beyond} incorporates a reward function to filter the dataset sampled from the current policy. The function provides binary rewards based on the correctness of the generated samples rather than a learned reward model trained on human preferences in ReST~\cite{gulcehre2023reinforced}. AutoAct~\cite{qiao2024autoact} leverages F1-score and accuracy as rewards for synthetic trajectories and collects trajectories with exactly correct answers for further training. 
Self-Talk~\cite{ulmer2024bootstrapping} measures the number of completed subgoals to filter the generated dialogues, ensuring that only high-quality data is used for training. To encourage diversity of the source instructions, Self-Instruct~\cite{wang2023self} automatically filters low-quality or repeated instructions using ROUGE-L similarity and heuristics before adding them to the task pool. 

The filtering criteria or metrics are crucial for maintaining the quality and reliability of the generated outputs, thereby ensuring the continuous improvement of the model's capability. 


\subsubsection{Metric-Free}
Some methods seek filtering strategies beyond external metrics, making the process more flexible and adaptable. \textit{Metric-Free} filtering typically involves sampling outputs and evaluating them based on internal consistency measures or other model-inherent criteria~\cite{huang2022large, weng2023self, chen2022codet}. The filtering in Self-Consistency~\cite{wang2023SC} is based on the consistency of the final answer across multiple generated reasoning paths, with higher agreement indicating higher reliability. LMSI~\cite{huang2022large} utilizes CoT prompting plus self-consistency for generating high-confidence self-training data.

Designing internal consistency measures that accurately reflect output quality can be challenging. Self-Verification~\cite{weng2023self} allows the model to select the candidate answer with the highest interpretable verification score, calculated by assessing the consistency between the predicted and original condition values. For the code generation task, CodeT~\cite{chen2022codet} considers both the consistency of the outputs against the generated test cases and the agreement of the outputs with other code samples. 

These methods emphasize the language model's ability to self-assess and filter its outputs based on internal agreement, showcasing a significant step forward in self-evolution without the direct intervention of external metrics.

\subsection{Correcting}
Recent advancements emphasize the importance of iterative self-correction in LLMs, allowing for the refinement of experiences.
This section divides the methods employed into two categories: \textit{Critique-Based} and \textit{Critique-Free} correction. Critiques often serve as strong hints that include the rationale behind perceived errors or suboptimal outputs, guiding the model towards improved iterations.

\subsubsection{Critique-Based}
These methods rely on additional judging processes to draw the critiques of the experiences. Then, the experiences are refined based on the critiques.
By leveraging either self-generated~\cite{mad2023refine, bai2022constitutional, shinn2023reflexion, lu2023self} or environment-interaction generated critiques~\cite{gou2023critic, jiang2023selfevolve, zhou2023isr}, the model benefits from detailed feedback for nuanced correction. 

LLMs have demonstrated their ability to identify errors in their outputs. Self-Refine~\cite{mad2023refine} introduces an iterative process in which the model refines its initial outputs conditioned on actionable self-feedback without additional training. To evolve from the correction, CAI~\cite{bai2022constitutional} generates critiques and revisions of its outputs in the supervised learning phase, significantly improving the initial model. Applied to an agent automating computer tasks, RCI~\cite{kim2023rci} improves its previous outputs based on the critique finding errors in the outputs. 

Since weaker models may struggle significantly with the self-critique process, several approaches enable models to correct the outputs using critiques provided by external tools. CRITIC~\cite{gou2023critic} allows LLMs to revise the output based on the critiques obtained during interaction with tools in general domains. SelfEvolve~\cite{jiang2023selfevolve} prompts an LLM to refine the answer code based on the error information thrown by the interpreter. ISR-LLM \cite{zhou2023isr} helps the LLM planner find a revised action plan by using a validator in an iterative self-refinement process.

The primary advantage of this method lies in its ability to process and react to detailed feedback, potentially leading to more targeted and nuanced corrections. 

\subsubsection{Critique-Free}
Contrary to critique-based, critique-free methods correct the experiences directly leveraging objective information~\cite{zelikman2022star, chen2023teaching, chen2023iter, gero2023self}. These methods offer the advantage of independence from nuanced feedback that critiques provide, allowing for corrections that adhere strictly to factual accuracy or specific guidelines without the potential bias introduced by critiques. 

One group of critique-free methods modifies the experiences on the signal of whether the task was correctly resolved. 
Self-Taught Reasoner (STaR)~\cite{zelikman2022star} proposes a technique that iteratively generates rationales to answer questions. If the answers are incorrect, the model is prompted again with the correct answer to generate a more informed rationale. Self-Debug~\cite{chen2023teaching} enables the model to perform debugging steps by investigating execution results from unit tests and explaining the code on its own.

Different from depending on the task-solving signal, other information produced during the solving process can be leveraged. IterRefinement~\cite{chen2023iter} relies on a series of refined prompts that encourage the model to reconsider and improve upon its previous outputs without any direct critique. For information extraction tasks, Clinical SV~\cite{gero2023self} grounds each element in evidence from the input and prunes inaccurate elements using supplied evidence.

These critique-free approaches simplify the correction mechanism, allowing for easier implementation and faster adjustments.

\begin{figure}[!tb]
    \centering
    \includegraphics[width=1\columnwidth]{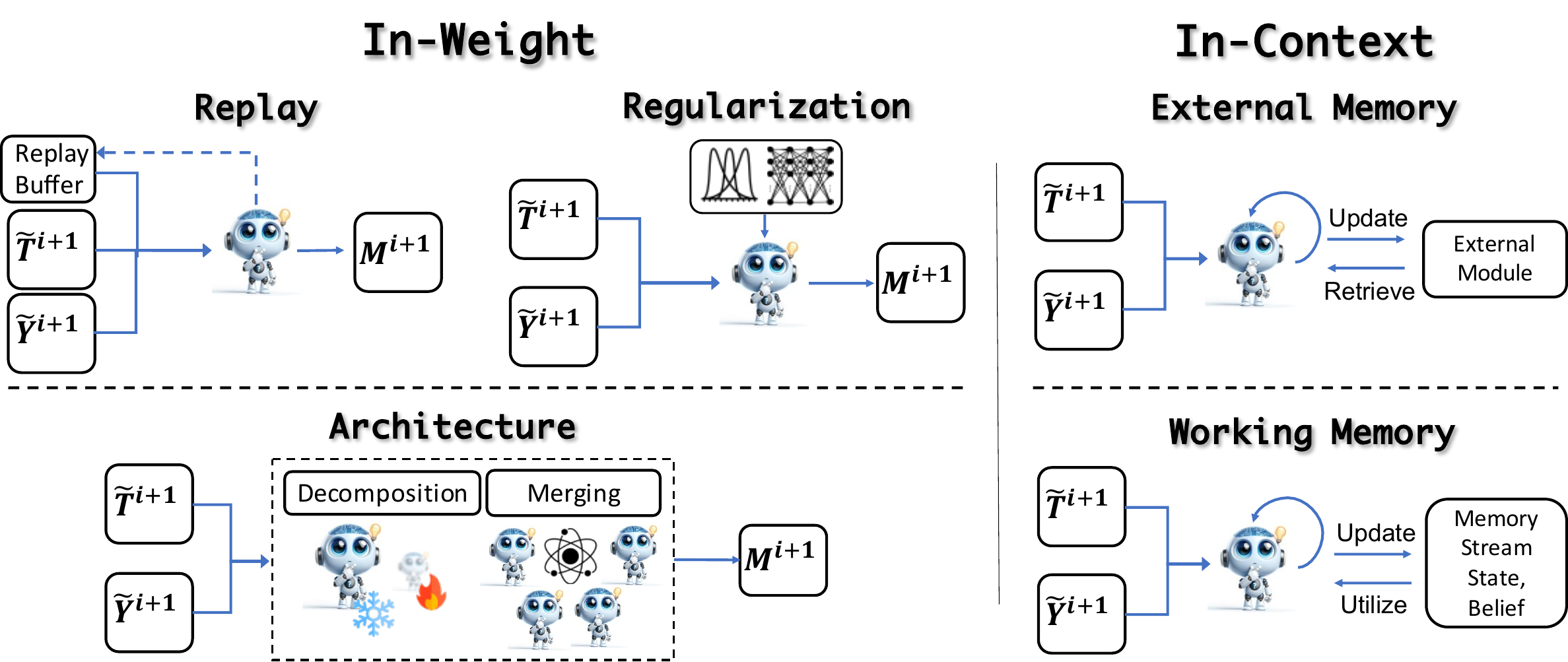}
    \caption{The illustration of \textit{Updating} methods, including in-weight and in-context updating. The terms $\Tilde{\gT}^{t}$ and $\Tilde{\gY}^{t}$ represent refined experiences, each containing task and corresponding solutions, respectively. $\mathrm{M}^{t}$ denotes the updated model. }
    \label{fig:updating}
\end{figure}

\section{Updating}
\label{sec:updating}
After experience refinement, we enter the crucial updating phase that leverages the refined experiences to improve model performance. We formulate updating as follows:
\begin{equation}
    \mathrm{M}^{t+1} = f^{\gU}(\Tilde{\gT}^{t}, \Tilde{\gY}^{t}, \gE^{t}, \mathrm{M}^{t}),
\end{equation}
where $f^{\gU}$ is the updating functions. These update methods keep the model effective by adapting to new experiences and continuously improving performance in changing environments and during iterative training.

We divide these approaches into in-weight learning, which involves updates to model weights, and in-context learning, which involves updates to external or working memory. 

\subsection{In-Weight}
Classical training paradigms in updating LLMs in weight encompass continuous pretraining \cite{brown2020language, roziere2023code}, supervised fine-tuning \cite{longpre2023flan}, and preference alignment \cite{ouyang2022training, touvron2023llama}. However, in the iterative training process of self-evolving, the core challenge lies in \textbf{achieving overall improvement} and \textbf{preventing catastrophic forgetting}, which entails refining or acquiring new capabilities while preserving original skills. Solutions to this challenge can be categorized into three main strategies: replay-based, regularization-based, and architecture-based methods.

\subsubsection{Replay-based} 
Replay-based methods reintroduce previous data to retain old knowledge. One is experience replay, which mixes the original and new training data to update LLMs \cite{roziere2023code, yang2024react, zheng2023self, lee2024llm2llm, wang2023adapting}. For example,  Reinforced Self-Training (ReST) \cite{gulcehre2023reinforced, aksitov2023rest} method iteratively updates large language models by mixing seed training data with filtered new outputs generated by the model itself. AMIE \cite{tu2024towards} utilizes a self-play simulated learning environment for iterative improvement and mixes generated dialogues with supervised fine-tuning data through inner and outer self-play loops. SOTOPIA-$\pi$ \cite{wang2024sotopia}  leverages behavior cloning from the expert model and self-generated social interaction trajectory to reinforce positive behaviors.

Another is generative replay, which adopts the self-generated synthesized data as knowledge to mitigate catastrophic forgetting. For instance, Self-Synthesized Rehearsal (SSR) \cite{huang2024mitigating} generates synthetic training instances for rehearsal, enabling the model to preserve its ability without relying on real data from previous training stages. Self-Distillation Fine-Tuning (SDFT) \cite{yang2024self} generates a distilled dataset from the model itself to bridge the distribution gap between task datasets and the LLM's original distribution to mitigate catastrophic forgetting.

\subsubsection{Regularization-based}
Regularization-based methods constrain the model's updates to prevent significant deviations from original behaviors, exemplified by function- and weight-based regularization. Function-based regularization focuses on modifying the loss function that a model optimizes during training \cite{zhong2023self, peng2023token}. For example, InstuctGPT \cite{ouyang2022training} employs a per-token KL-divergence penalty from the output probabilities of the initial policy model $\pi_\text{SFT}$ on the updated policy model $\pi_\text{RL}$. FuseLLM \cite{wan2024knowledge} employs a technique akin to knowledge distillation \cite{hinton2015distilling}, leveraging the generated probability distributions from source LLMs to transfer the collective knowledge to the target LLM.

Weight-based regularization \cite{kirkpatrick2017overcoming} directly targets the model's weights during training. Techniques such as Elastic Reset \cite{noukhovitch2024language} counters alignment drift in RLHF by periodically resetting the online model to an exponentially moving average of its previous states. Furthermore, \citet{rame2024warm} introduced WARM, which combines multiple reward models through weight averaging to address reward hacking and misalignment. Moreover, AMA \cite{lin2024mitigating} adaptively average model weights to optimize the trade-off between reward maximization and forgetting mitigation.
    
\subsubsection{Architecture-based}
Architecture-based methods explicitly utilize extra parameters or models for updating, including decomposition- and merging-based approaches.
Decomposition-based methods separate large neural network parameters into general and task-specific components and only update the task-specific parameters to mitigate forgetting. LoRA \cite{hu2021lora, dettmers2024qlora} inject trainable low-rank matrices to significantly reduce the number of trainable parameters while maintaining or improving model performance across various tasks. This paradigm is later adopted by GPT4tools \cite{yang2024gpt4tools}, OpenAGI \cite{ge2024openagi} and Dromedary \cite{sun2024principle}. Dynamic ConPET \cite{song2023conpet} combines pre-selection and prediction with task-specific LoRA modules to prevent forgetting, ensuring scalable and effective adaptation of LLMs to new tasks.

Merging-based methods, on the other hand, involve combining multiple models or layers to achieve general improvements, including but not limited to merging multiple generic and specialized model weights into a single model \cite{wortsman2022model, ilharco2022editing, yu2023language, yadav2024ties}, through mixture-of-expert approach \cite{ding2024mastering} or even layer-wise merging and up-scaling such as EvoLLM \cite{akiba2024evolutionary}.

\subsection{In-Context}
In addition to directly updating model parameters, another approach is to leverage the in-context capabilities of LLMs to learn from experiences, thereby enabling fast adaptive updates without expensive training costs. The methods could be divided into updating external and working memory.

\begin{table}
\centering 
\setlength{\tabcolsep}{3mm}
\begin{tabular}{ccc}
\toprule 
\textsc{Method} & \textsc{Content} & \textsc{Operation} \\
\midrule
MoT~\cite{li2023mot} 
& \texttt{Experience} & \texttt{Insert} \\
TRAN~\cite{yang2023failures}
& \texttt{Rationale} & \texttt{Insert}, \texttt{Reflect}  \\
MemoryBank~\cite{zhong2024memorybank}
& \texttt{Experience}, \texttt{Rationale} & \texttt{Insert}, \texttt{Reflect}, \texttt{Forget} \\
MemGPT~\cite{packer2023memgpt} 
& \texttt{Experience} & \texttt{Insert}, \texttt{Forget} \\
TiM~\cite{liu2023think} 
& \texttt{Rationale} & \texttt{Insert} \\
IML~\cite{wang2024memory} 
& \texttt{Rationale} & \texttt{Insert}, \texttt{Reflect} \\
ICE~\cite{qian2024investigate}
& \texttt{Rationale} & \texttt{Insert}, \texttt{Reflect} \\
AesopAgent~\cite{wang2024aesopagent}
& \texttt{Experience}, \texttt{Rationale} & \texttt{Insert}, \texttt{Reflect} \\
\bottomrule 
\end{tabular}
\caption{Content and operations of updating external memory.} 
\label{tab:memory} 
\end{table}

\subsubsection{External Memory} 
This approach utilizes an external module to collect, update, and retrieve past experiences and knowledge, enabling models to access a rich pool of insights and achieve better results without updating model parameters. The external memory mechanism is common in AI Agent systems~\cite{xu2023exploring, qian2024investigate, wang2024aesopagent}.
This section provides a detailed overview of the latest methods for updating external memory, emphasizing the aspects of memory \textit{Content} and \textit{Updating Operations}, and summarized in Table~\ref{tab:memory}.

\textit{Content:} External memory mainly stores two types of content: past experiences and reflected rationale, each serving distinct purposes. For instance, past experience provides valuable historical context, serving as a guiding force toward achieving improved outcomes.
MoT~\cite{li2023mot} archives filtered question-answer pairs to construct a beneficial memory repository. 
Additionally, the FIFO Queue mechanism in MemGPT~\cite{packer2023memgpt} maintains a rolling history of messages, encapsulating interactions between agents and users, system notifications, and inputs and outputs of function calls.

On the other hand, reflected rationales offer condensed explanations, such as rules, that support decision-making. For instance, 
TRAN~\cite{yang2023failures} archives rules inferred from experiences alongside information on mistakes to mitigate future errors.
Correspondingly, TiM~\cite{liu2023think} preserves inductive reasoning, defined as text elucidating the relationships between entities.
Moreover, IML~\cite{wang2024memory} and ICE~\cite{qian2024investigate} store comprehensive notes and rules derived from a series of trajectories, demonstrating the broad spectrum of content types that memory systems can accommodate.

MemoryBank~\cite{zhong2024memorybank} and AesopAgent~\cite{wang2024aesopagent} reflect on experience and rationale, save them as external memory, and retrieve them when needed to achieve better performance.

\textit{Updating Operation:} We categorize the operations to the memory into Insert, Reflect, and Forget. The most common operation is insert, methods insert text content into the memory for storage~\cite{li2023mot, yang2023failures, zhong2024memorybank, packer2023memgpt, liu2023think, wang2024memory}.
Another operation is reflection, which is to think and summarize previous experiences to conceptualize rules and knowledge for future use~\cite{yang2023failures, zhong2024memorybank, wang2024memory, qian2024investigate}.
Last, due to the limited storage of memory, forgetting content is crucial to keeping memory efficient and the content valid. 
MemGPT~\cite{packer2023memgpt} adopts the FIFO queue to forget the contents.
MemoryBank~\cite{zhong2024memorybank} establishes a forgetting curve on the insert time of each item.

\subsubsection{Working Memory} The methods use past experience to evolve the capabilities of agents by updating internal memory streams, states, or beliefs, known as working memory, often in the form of verbal cues. Reflexion~\cite{shinn2023reflexion} introduces verbal reinforcement learning for decision-making improvement without conventional model updates. Similarly, IML~\cite{wang2024memory} enables LLM-based agents to autonomously learn and adapt to their environment by summarizing, refining, and updating knowledge based on past experience directly in working memory. 

EvolutionaryAgent~\cite{li2024agent} aligns agents with dynamically changing social norms through evolution and selection principles, leveraging environmental feedback for self-evolution. Agent-Pro~\cite{zhang2024agent} employs policy-level reflection and optimization, allowing agents to adapt their behavior and beliefs in interactive scenarios based on past outcomes. Lastly, ProAgent~\cite{zhang2024proagent} enhances cooperation in multi-agent systems by dynamically interpreting teammates' intentions and adapting behavior. 

These collective works demonstrate the importance of integrating past experiences and knowledge into the agents' memory stream to refine their state or beliefs for improved performance and adaptability across various tasks and environments.

\section{Evaluation}
\label{sec:evaluation}
Much like the human learning process, it is essential to ascertain whether the present level of ability is adequate and meets the application requirements through evaluation. Furthermore, it is from these evaluations that one can identify the direction for future learning.
However, how to accurately assess the performance of an evolved model and provide directions for future improvements is a crucial yet underexplored research area. For a given evolved model $M^t$, we conceptualize the evaluation process as follows:
\begin{equation}
  \gE^{t+1}, \gS^{t} = f^{\gE}(M^t, \gE^{t}, \mathrm{ENV}),
\end{equation}
where $f^E$ represents the evaluation function that measures the performance score ($\gS^{t+1}$) of the current model and provide evolving goal ($\gE^{t+1}$) for the next iteration. Evaluation function $f^\gE$ can be categorized into quantitative and qualitative approaches, each providing valuable insights into model performance and areas for improvement.

\subsection{Quantitative Evaluation}
This method focuses on providing measurable metrics to reliably assess LLM performance, such as automatic \cite{papineni2002bleu, lin2004rouge} and human evaluation. However, traditional automatic metrics struggle to accurately evaluate increasingly complex tasks, and human assessment is not an ideal option for autonomous self-evolution. Recent trends use LLMs as human proxy for automatic evaluators, offering cost-effective and scalable solutions for evaluations. 

For example, reward model score has been widely used to measure model or task performances \cite{shinn2023reflexion} and select the best checkpoint \cite{ouyang2022training}. LLM-as-a-judge \cite{zheng2024judging} using LLMs to evaluate LLMs, employing methods like pairwise comparison, single answer grading, and reference-guided grading. It shows that LLMs can closely match human judgment, enabling efficient large-scale evaluations.

\subsection{Qualitative Evaluation}
Qualitative evaluation involves case studies and analysis to derive insights, offering evolving guidance for subsequent iterations. Initiatives such as LLM-as-a-judge \cite{zheng2024judging} provide the reasoning behind its assessments; ChatEval \cite{chan2023chateval} explores the strengths and weaknesses of model outputs through debate mechanisms. Furthermore, TRAN \cite{yang2023failures} leverages past errors to formulate rules that enhance future LLM performances. Nonetheless, compared with instance-level critic or reflection, qualitative evaluation at the task- or model-level still needs comprehensive investigation.

\section{Open Problems}
\label{sec:future}

\subsection{Objectives: Diversity and Hierarchy}
\label{sec:objective coverage}
Section~\ref{sec:objective} summarizes existing evolution objectives and their coverage. Nonetheless, these highlighted objectives can only satisfy a small fraction of the vast human needs. 
The extensive application of LLMs across various tasks and industries highlights unresolved challenges in establishing self-evolution frameworks for evolving objectives that can comprehensively address a broader spectrum of real-world tasks~\cite{eloundou2023gpts}. 

Furthermore, the concept of evolving objectives entails a potential hierarchical structure; for instance, UltraTool~\cite{huang2024planning} and T-Eval~\cite{chen2023t} categorize the capability of tool usage into various sub-dimensions. Exploring evolutionary objectives into manageable sub-goals and pursuing them individually emerges as a viable strategy. 

Overall, a clear and urgent need exists to develop self-evolution frameworks that effectively address diversified and hierarchical objectives.

\subsection{Level of Autonomy: From Low to High}
\label{sec:autonomous level}
Self-evolution in large models is emerging, yet lacks clear definitions for its autonomous levels. We categorize self-evolution into three tiers: low, medium, and high-level autonomy
\noindent\paragraph{Low-level} In this level, the user predefined the evolving object $\gE$ and it remains unchanged. The user needs to design the evolving pipeline, namely all modules $f^{\bullet}$, on its own. Then, the model completes the self-evolution process based on the designed framework. We denote this level of self-evolution in the following formula:
\begin{equation}
    \mathrm{\Tilde{M}} = \mathrm{Evol^{L}}(\mathrm{M}, \gE, f^{\bullet}, \mathrm{ENV}),
\end{equation}
where $\mathrm{M}$ denotes the model to be evolved. $\mathrm{\Tilde{M}}$ is the evolving output. $\mathrm{ENV}$ is the environment. Most of the current works lie at this level.

\noindent\paragraph{Medium-level} In this level, the user only sets the evolving object $\gE$ and keeps it unchanged. 
The user doesn't need to design the specific modules $f^{\bullet}$ in the framework. The model can construct each module $f^{\bullet}$ independently for self-evolution. This level denotes as follows:
\begin{equation}
    \mathrm{\Tilde{M}} = \mathrm{Evol^{M}}(\mathrm{M}, \gE, \mathrm{ENV}),
\end{equation}

\noindent\paragraph{High-level} In the final level, the model diagnoses its deficiency and constructs the self-evolution methods to improve itself. This is the ultimate purpose of self-evolution. The user model sets its own evolving object $\gE$ according to the evaluation $f^{E}$ output. The evolving objective would change during the iteration. 
Besides, the model designs the specific modules $f^{\bullet}$ in the framework. 
We represent this level as:
\begin{equation}
    \mathrm{\Tilde{M}} = \mathrm{Evol^{H}}(\mathrm{M}, \mathrm{ENV}),
\end{equation}

As discussed in the previous open problem (\S~\ref{sec:objective coverage}), there are a large of unfulfilled objectives. 
However, most of the existing self-evolution frameworks are at the Low-level which requires specifically designed modules~\cite{yuan2024self, lu2024large, qiao2024autoact}. These frameworks are objective-dependent and rely on large human efforts to develop. 
Exhausting all objectives are not deployment-efficient which brings about the urgent need to develop medium and high levels self-evolution frameworks.
At the medium level, it doesn't require expert efforts to design specific modules. LLMs can self-evolve according to targeted objectives.
Then at the high level, LLMs can investigate their current deficiencies and evolve in a targeted manner.
In all, developing highly autonomous self-evolution frameworks remains an open problem.

\subsection{Experience Acquisition and Refinement: From Empirical to Theoretical} 
Suppose we have addressed the previous two challenges and developed promising self-evolution frameworks, but the exploration of self-evolution LLMs still lacks solid theoretical grounding. This idea posits that LLMs can self-improve or correct their outputs, with or without feedback from the environment. However, the mechanisms behind it remain unclear. Studies show mixed results: \citet{huang2023large} observed self-corrective behavior in models with over 22 billion parameters, while \citet{ganguli2023capacity} finds LLMs struggle to self-correct reasoning errors without external feedback.

A related challenge is the use of self-generated data for learning. Critics argue this approach could reduce linguistic diversity \cite{guo2023curious} and lead to "model collapse," where models fail to capture complex, long-tailed data distributions \cite{shumailov2023curse}. Furthermore, \citet{alemohammad2023self} reveal that generative models trained on their synthetic outputs progressively lose output quality and diversity. \citet{fu2024towards} extend this by theoretically analyzing the impact of self-consuming training loops on model performance, emphasizing the importance of balancing synthetic and real data to mitigate error accumulation. 

Recent studies \cite{yang2024react, singh2023beyond} also show that current methods struggle to improve after more than three rounds of self-evolution. One hypothesized reason is that the self-critic of LLM has not co-evolved with the evolving objective, but more experimental and theoretical support is still needed. These findings highlight a pressing need for more theoretical exploration in self-evolving LLMs. Addressing these concerns is crucial for advancing the field and ensuring that models can effectively learn and improve over time.

\subsection{Updating: Stability-Plasticity Dilemma} 
The stability-plasticity dilemma represents a crucial yet unresolved challenge that is essential for iterative self-evolution. This dilemma reflects the difficulty of balancing the need to retain previously learned information (stability) while adapting to new data or tasks (plasticity). Existing LLMs either overlook this issue or adopt conventional methods that may be ineffective. While training models from scratch could mitigate the problem of catastrophic forgetting, it is highly inefficient, particularly as model parameters increase exponentially and autonomous learning capabilities advance. Finding a balance between acquiring new skills and preserving existing knowledge is crucial for achieving effective and efficient self-evolution, leading to overall improvement.

\subsection{Evaluation: Systematic and Evolving}
To effectively assess LLMs, a dynamic, comprehensive benchmark is crucial. This becomes even more pivotal as we progress towards Artificial General Intelligence (AGI). Traditional static benchmarks risk obsolescence due to LLMs' evolving nature and potential access to test data through interacting with environments, such as search engines, undermining their reliability. A dynamic benchmark, like Sotopia \cite{zhou2023sotopia}, proposes a solution by creating an LLM-based environment tailored for evaluating the social intelligence of LLMs, thereby avoiding the limitations posed by static benchmarks.

\subsection{Safety and Superalignment}
The advancement of LLMs opens the possibility for AI systems to achieve or even surpass expert-level capabilities in both supportive and autonomous decision-making. For safety, ensuring these LLMs align with human values and preferences is crucial, particularly to mitigate inherent biases that can impact areas such as political debates, as highlighted by \citet{taubenfeld2024systematic}.  OpenAI's initiative, Superalignment \cite{superalignment2023}, aims to align a superintelligence by developing scalable training methods, validating models for alignment, and stress-testing the alignment process through scalable oversight \cite{saunders2022self}, robustness \cite{perez2022red}, automated interpretability \cite{bills2023language}, and adversarial testing. Although challenges remain, Superalignment marks an initial attempt to develop a self-evolving LLM that closely aligns with human ethics and values in a scalable way.





\section{Conclusion}
\label{sec:conclusion}
The evolution of LLMs towards self-evolution paradigms represents a transformative shift in artificial intelligence akin to the human learning process. It is promising to overcome the limitations of current models that rely heavily on human annotation and teacher models. 
This survey presents a comprehensive framework for understanding and developing self-evolving LLMs, structured around iterative cycles of experience acquisition, refinement, updating, and evaluation. By detailing advancements and categorizing the evolution objectives within this framework, we offer a thorough overview of current methods and highlight the potential for LLMs to adapt, learn, and improve autonomously. 
We also identify existing challenges and propose directions for future research, aiming to accelerate the progress toward more dynamic, intelligent, and efficient models. This work deepens the understanding of self-evolving LLMs. It paves the way for significant advancements in AI, marking a step towards achieving superintelligent systems capable of surpassing human performance in complex real-world tasks.


\begin{appendices}

\end{appendices}

\end{document}